\documentclass{article}
\usepackage{graphicx}

\PassOptionsToPackage{numbers,sort}{natbib}
\usepackage[main, final]{neurips_2026}

\usepackage[utf8]{inputenc} 
\usepackage[T1]{fontenc}    
\usepackage{hyperref}       
\usepackage{url}            
\usepackage{booktabs}       
\usepackage{amsfonts}       
\usepackage{nicefrac}       
\usepackage{microtype}      
\usepackage{xcolor}         
\usepackage{makecell}

\usepackage{graphicx}
\usepackage{multirow}
\usepackage{subfigure}
\usepackage{enumitem}
\usepackage{amsmath,amssymb,amsfonts}

\title{SIGMA: Semantic-Difference Instruction-Grounding Mask Annotator for Text-Driven Image Manipulation Localization}

\author{%
  \textbf{Peiyu Zhuang}\textsuperscript{1},
  \textbf{Jianquan Yang}\textsuperscript{1}\thanks{Corresponding author.},
  \textbf{Haodong Li}\textsuperscript{2},
  \textbf{Zhuoying Cai}\textsuperscript{1},
  \textbf{Ruitao Xie}\textsuperscript{3},
  \textbf{Jishen Zeng}\textsuperscript{4},\\
  \textbf{Baoying Chen}\textsuperscript{4},
  \textbf{Jiwu Huang}\textsuperscript{5},
  \textbf{Xiaochun Cao}\textsuperscript{1}\footnotemark[1]
  \\[2pt]
  \textsuperscript{1}Shenzhen Campus of Sun Yat-sen University, China
  \\
\textsuperscript{2}Shenzhen University, China
  \\[-1pt]
  \textsuperscript{3}Shenzhen University of Advanced Technology, China
  \\
  \textsuperscript{4}Alibaba Group, China
  \quad
  \textsuperscript{5}Shenzhen MSU-BIT University, China
  \\[2pt]
  \texttt{\{zhuangpy, yangjq65, caoxiaochu\}@mail.sysu.edu.cn;}
  \\[-1pt]
  \texttt{lihaodong@szu.edu.cn; caizhy55@mail2.sysu.edu.cn}
  \\[-1pt]
  \texttt{xieruitao@suat-sz.edu.cn; jishen.zjs@alibaba-inc.com;}
  \\[-1pt]
  \texttt{1900271059@email.szu.edu.cn;jwhuang@smbu.edu.cn}
}

\begin{document}

\maketitle

\begin{abstract}
Text-driven image editing has advanced rapidly, but reliably localizing these manipulations requires image manipulation localization (IML) models trained on large pixel-annotated datasets, and there is still no low-cost way to obtain such training data at scale.
We observe that these data already exist in disguise: public editing datasets contain millions of structurally identical $\textit{(original, edited)}$ pairs to IML training samples, lacking only pixel-level masks. Recovering these masks automatically is non-trivial: pixel differencing is overwhelmed by diffusion-induced perturbations across all pixels, and instruction-only grounding localizes only what the prompt describes, missing unintended editor side-effects.
We propose $\textbf{SIGMA}$ ($\textbf{S}$emantic-difference $\textbf{I}$nstruction-$\textbf{G}$rounding $\textbf{M}$ask $\textbf{A}$nnotator), which performs semantic-feature differencing in a vision foundation backbone and injects an instruction-derived spatial prior into this visual stream via bidirectional cross-modal refinement, amplifying the difference signal at intended-edit regions when the editor faithfully realizes user intent. SIGMA is trained in two complementary stages: Stage I supervises on inpainting masks and Stage II closes the diffusion-domain shift via VAE-roundtrip noise calibration, EMA self-training, and an edit-noise disentanglement loss. SIGMA outperforms existing automatic mask generators on five benchmarks ($\textbf{+12.20\%}$ F1, $\textbf{+11.16\%}$ IoU). When applied to public editing corpora, it produces a $\sim$1.1M IML training set that improves four diverse detectors by $\textbf{+20.98\%}$ F1 across five datasets, turning previously unused editing data into a model-agnostic supervisory resource for IML. For reproducibility, our project page is available at \url{https://zhuangpeiyu.github.io/projects/SIGMA/}.
\end{abstract}

\section{Introduction}
\label{sec:intro}

Recent advances in text-driven image editing~\cite{brooks_InstructPix2PixLearningFollow_2023, hertz_PrompttoPromptImageEditing_2022, liu_Step1XEditPracticalFramework_2025, labs_FLUX1KontextFlow_2025, wu_OmniGen2ExplorationAdvanced_2025, wu_QwenImageTechnicalReport_2025} have lowered the barrier to producing semantically coherent forgeries that fuel misinformation and erode public trust. Mitigating this risk demands robust image manipulation localization (IML) models, which in turn require large-scale, pixel-accurately annotated training data, a resource the IML community currently lacks. Nowadays, public editing corpora have reached the million-sample scale (\emph{e.g.}, ByteMorph\cite{chang2025bytemorph} 6.0M, AnyEdit\cite{yu_AnyEditMasteringUnified_2025} 2.5M), while the largest IML resources with pixel-level masks (SIDA\cite{huang2025sida} 300K, GRE\cite{sun2024rethinking} 228.7K) are dwarfed by even a single editing corpus (see Fig.~\ref{fig:dataset_scale} in Appendix~\ref{sec:dataset_scale} for additional information). We note that existing editing datasets already comprise \textit{(original, edited)} pairs that are structurally equivalent to the training samples used in IML; however, they lack the corresponding pixel-level annotation masks. 
This raises a natural question: \emph{can existing editing pairs be automatically turned into pixel-accurate IML supervision, enabling million-scale corpora without rerunning editors or manual annotation?}

\begin{figure}[t]
    \centering
    \includegraphics[width=\textwidth]{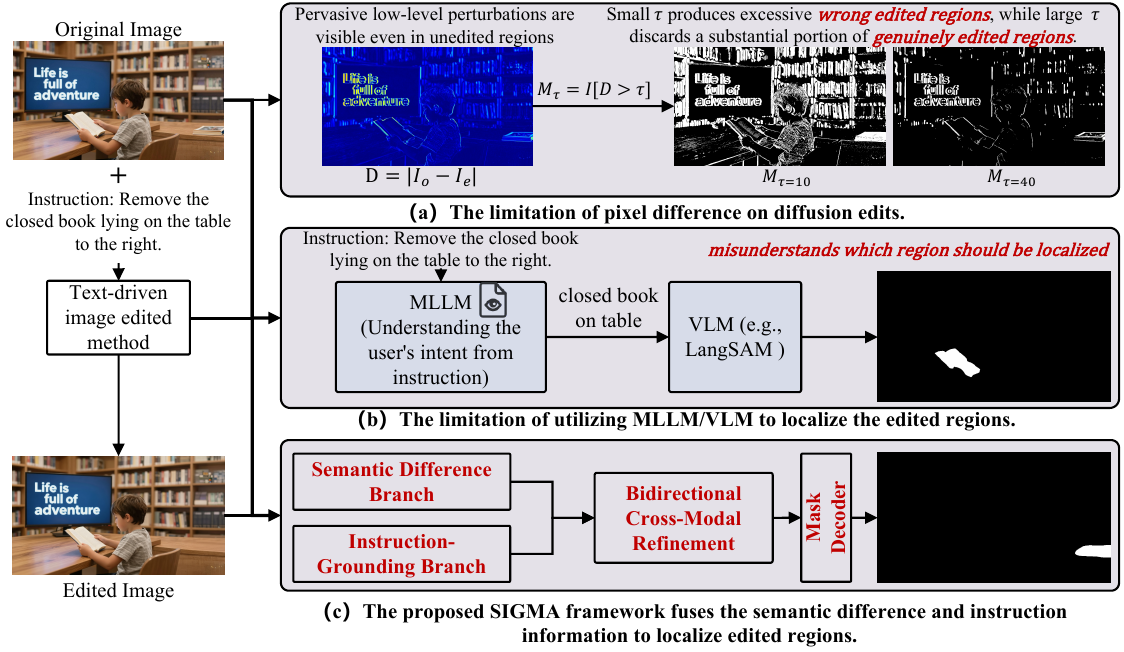}
    \caption{(a) Failure modes of 
    pixel-level differencing under diffusion-based text-driven editing. 
    (b) Failure modes of MLLM/VLM-based mask inference from instructions. 
    (c) Overview of the SIGMA framework.}
    \label{fig:motivation}
\end{figure}

Answering this question hinges on an automatic, scalable annotator. Two seemingly natural strategies prove ineffective. \textbf{Pixel-level differencing with thresholding} works for inpainting-style edits but breaks down under diffusion-based editing, where every pixel is perturbed by VAE encoding/decoding and iterative denoising. As shown in Fig.~\ref{fig:motivation}(a), low thresholds drown the mask in reconstruction noise, while high thresholds discard subtle yet semantically meaningful edits. \textbf{Instruction-grounding via MLLMs/VLMs} (Fig.~\ref{fig:motivation}(b)) localizes only what the prompt requests and cannot verify whether the editor faithfully realized that intent or introduced unintended side-effects. Hybrid pipelines (e.g., SAM with human refinement) yield accurate masks but trade scalability for accuracy. The two automatic strategies are not coincidentally inadequate: they are \emph{complementary, not alternative}. Pixel differencing observes what changed but cannot separate noise from edits, while instruction grounding encodes what was intended but cannot verify execution. 

Building on this view, we propose $\textbf{SIGMA}$ ($\textbf{S}$emantic-difference $\textbf{I}$nstruction-$\textbf{G}$rounding $\textbf{M}$ask $\textbf{A}$nnotator), a two-branch framework whose semantic branch performs multi-scale differencing on dense features from a frozen DINOv2~\cite{oquab_DINOv2LearningRobust_2024} encoder (\emph{where content actually differs}), while its instruction branch grounds a parsed transformation tuple (original concept, edited concept, action) onto each image via a vision-language grounder (\emph{where the change was intended}). A bidirectional cross-modal refinement module then injects the instruction prior into the visual stream to amplify true edit signals while continuously re-anchoring the prior to the most supported regions. To bridge supervision scarcity and diffusion-domain shift, SIGMA is trained in two stages: \textbf{Stage~I} pre-trains on inpainting pairs with exact masks, and \textbf{Stage~II} adapts to unlabeled editing pairs via VAE-roundtrip zero-edit calibration, exponential moving average (EMA) self-training, and an edit-noise feature disentanglement loss, jointly suppressing diffusion-induced false positives without any target-domain annotation.

Our contributions can be summarized as follows.
\begin{itemize}[leftmargin=*, labelsep=0.5em, itemindent=0pt, listparindent=0pt]
    \item We reframe scalable IML annotation as \textbf{semantic change localization}, showing that pixel differencing and instruction grounding are complementary but individually insufficient for diffusion-based edits.
    \item We propose \textbf{SIGMA}, a two-branch framework fusing semantic-feature differencing with parsed-instruction grounding via bidirectional cross-modal refinement, trained by a two-stage strategy that targets supervision scarcity and domain shift respectively.
    \item Using SIGMA, we construct a \textbf{million-scale IML dataset} from public editing corpora at no re-generation or manual annotation cost, achieving \textbf{+12.20\%} F1 over prior automatic mask generators and boosting four diverse IML detectors by \textbf{+20.98\%} F1 in cross-dataset generalization.
\end{itemize}

\section{Related Work}
\label{sec:related}

\textbf{IML and its data bottleneck.}
IML has progressed from hand-crafted forensic cues to learning-based methods exploiting noise inconsistencies, frequency artifacts, or diffusion priors~\cite{wu_ManTraNetManipulationTracing_2019, guillaro_TruForLeveragingAllRound_2023, guo_HierarchicalFineGrainedImage_2023a, yu_DiffForensicsLeveragingDiffusion_2024, su_CanWeGet_2025,huang2025sida,ZhuangDFCN, ZhuangReloc, li2025towards, yu2026toward, yu2025reinforced, yu2023learning}, with recent work integrating multimodal large language models for explainable, pixel-level localization~\cite{xu2025fakeshield, sun_ForgerySleuthEmpoweringMultimodal_2025a, kang_LEGIONLearningGround_2025}. Yet their performance is bottlenecked by training data: early benchmarks~\cite{Columbia, CASIA, novozamsky2020imd2020, mahfoudi2019defacto} rely on hand-crafted operations whose artifacts no longer reflect modern forgeries, while diffusion-era datasets~\cite{guillaro_TruForLeveragingAllRound_2023, jia_AutoSpliceTextPromptManipulated_2023, zhang_MagicBrushManuallyAnnotated_2023, zhang_DEAL300KDiffusionbasedEditing_2025a, wang_OpenSDISpottingDiffusionGenerated_2025} remain dwarfed by public text-driven editing corpora in scale and edit diversity (Fig.~\ref{fig:dataset_scale} in Appendix~\ref{sec:dataset_scale}). Our contribution is orthogonal to architectural advances: rather than proposing yet another localization model, we provide a scalable, fully automated annotator that turns previously unused text-driven editing data into supervision for arbitrary downstream IML models.
 
\textbf{Change detection.}
SIGMA's semantic differencing branch shares architectural intuition with change detection (CD) in remote sensing, where co-registered multi-temporal images are compared via Siamese CNNs, Transformers~\cite{bandara_TransformerBasedSiameseNetwork_2022}, or foundation models such as SAM~\cite{gao_CombiningSAMLimited_2025} and pretrained DDPMs~\cite{chamindabandara_DDPMCDDenoisingDiffusion_2025}. Despite this shared backbone, the two tasks differ at two levels. First, CD nuisances are low-level and semantically independent of content, so Siamese differencing can suppress them. However, in text-driven editing, the entire image is re-synthesized through a VAE and iterative denoising, yielding high-level, content-correlated perturbations at every pixel that feature normalization alone cannot remove. 
Second, mainstream CD operates on the image pair alone. In our setting, however, the editing instruction provides a strong spatial prior that, when the editor faithfully realizes user intent, amplifies the visual difference signal at the targeted regions.

\section{Proposed Method}
\label{sec:method}

\subsection{Problem Formulation}
\label{sec:formulation}

Given a source image $I^o \in \mathbb{R}^{H \times W \times 3}$, its text-driven edited counterpart $I^e \in \mathbb{R}^{H \times W \times 3}$, and the editing instruction $T$, we predict a binary mask $M \in \{0,1\}^{H \times W}$ delineating the semantically edited region. A pixel $(x,y)$ is considered edited iff its underlying semantic properties (object identity, material, color, or contextual role) have meaningfully changed:
\begin{equation}
    \small
    M(x,y) = \mathbf{1}\!\left[
        \mathcal{S}(I^o)_{(x,y)} \neq \mathcal{S}(I^e)_{(x,y)}
    \right],
\end{equation}
where $\mathcal{S}(\cdot)$ denotes a high-level semantic representation, so that VAE roundtrip noise and stochastic denoising are excluded. This definition imposes two requirements that neither pixel differencing nor instruction grounding alone satisfies (shown in Sec.~\ref{sec:intro}).
SIGMA addresses (i) by lifting comparison from pixels to a frozen DINOv2~\cite{oquab_DINOv2LearningRobust_2024} feature space, and (ii) by parsing the instruction into a structured spatial prior and injecting it into the visual differencing branch via bidirectional cross-modal refinement.

\subsection{The Proposed SIGMA Framework}
\label{sec:SIGMA}
\begin{figure}[tbp]
    \centering
    \includegraphics[width=\textwidth]{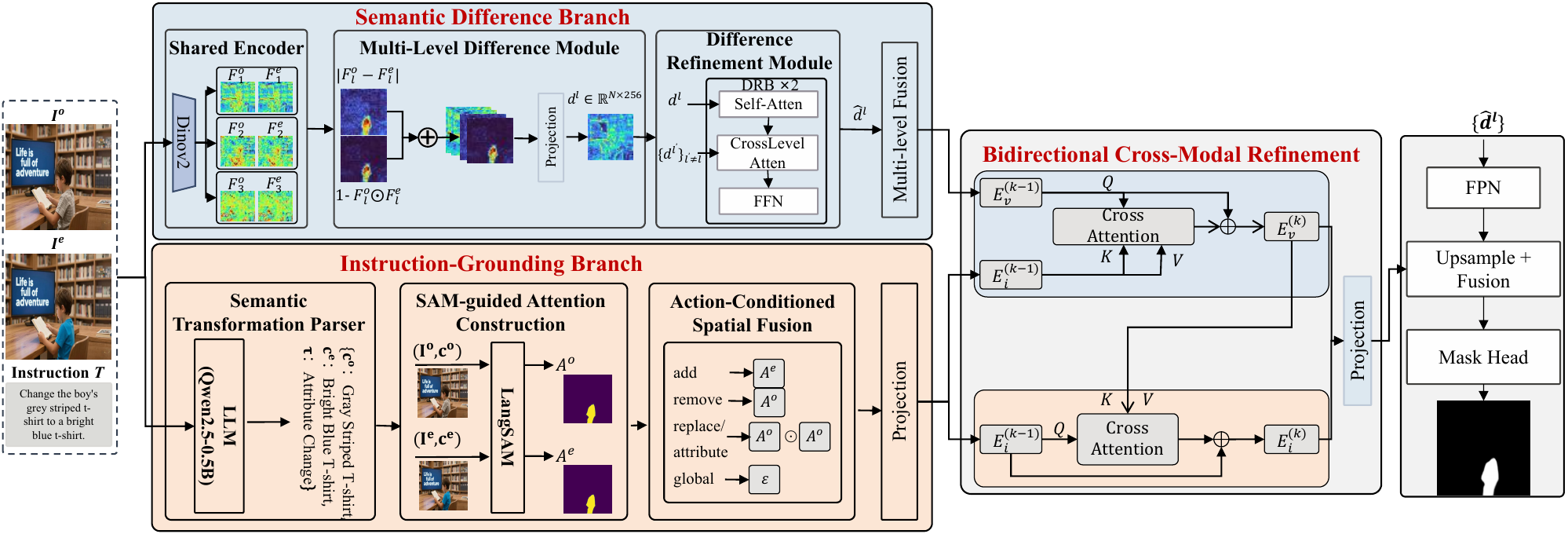}
    \caption{Overview of the SIGMA architecture.}
    \label{fig:proposedMethod}
\end{figure}

As illustrated in Fig.~\ref{fig:proposedMethod}, SIGMA is powered by four key components: a semantic difference branch, an instruction-grounding branch, a bidirectional cross-modal refinement module, and a mask decoder.

\subsubsection{Semantic-Difference Branch}
\textbf{Shared Encoder.} We adopt frozen DINOv2-Base as the shared backbone for both images. Both 
images are resized to $518\times518$, and patch token features are 
extracted at three designated layers $\mathcal{L} = \{l_1, l_2, l_3\}$ 
corresponding to layers 2, 5, and 11 of the ViT-Base architecture:
\begin{equation}
    \small 
    \mathbf{f}_o^l = \mathrm{DINOv2}(I^o),\quad
    \mathbf{f}_e^l = \mathrm{DINOv2}(I^e),\quad l \in \mathcal{L},
    \label{eq:dino_features}
\end{equation}
where $\mathbf{f}_o^l, \mathbf{f}_e^l \in \mathbb{R}^{N \times D}$, 
$N = 37\times37 = 1{,}369$, and $D = 768$. Shallow layers encode 
fine-grained texture and local structure, whereas deeper layers capture 
object-level semantics and holistic scene understanding. 

\textbf{Multi-Level Differencing Module.}
For each level $l \in \mathcal{L}$, we construct a difference 
representation that encodes both magnitude and directional dissimilarity 
of feature change. Let $\tilde{\mathbf{f}}_o^l$ and 
$\tilde{\mathbf{f}}_e^l$ denote the L2-normalized features. We compute
\begin{equation}
    \small
    \mathbf{d}_\mathrm{abs}^l 
        = \bigl|\tilde{\mathbf{f}}_o^l - \tilde{\mathbf{f}}_e^l\bigr|,
    \qquad
    \mathbf{d}_\mathrm{cos}^l 
        = \tilde{\mathbf{f}}_o^l \odot \tilde{\mathbf{f}}_e^l,
    \label{eq:d_abs_d_cos}
\end{equation}
We then fuse these signals with 
the original features and project into a shared embedding:
\begin{equation}
    \small
    \mathbf{d}^l = \mathrm{Proj}^l\!\Bigl(
        \bigl[\mathbf{d}_\mathrm{abs}^l \;\|\;
              \mathbf{d}_\mathrm{cos}^l \;\|\;
              \tilde{\mathbf{f}}_o^l \;\|\;
              \tilde{\mathbf{f}}_e^l\bigr]
    \Bigr),\quad
    \mathrm{Proj}^l\colon\mathbb{R}^{4D}\to\mathbb{R}^{C},\; C=256.
    \label{eq:dl_projection}
\end{equation}

\textbf{Difference Refinement Module.}
Raw difference features are corrupted by patch-level quantization 
artifacts. We refine them with transformer-based Difference Refinement 
Blocks (DRBs):
\begin{equation}
    \small
    \hat{\mathbf{d}}^l = \mathrm{DRB}^l\!\left(
        \mathbf{d}^l,\;\bigl\{\mathbf{d}^{l'}\bigr\}_{l'\neq l}
    \right).
    \label{eq:drb}
\end{equation}
Each DRB consists of a within-level self-attention layer, a cross-level 
attention layer (queries at $l$ attend to keys and values from all 
$l'\neq l$), and a feed-forward network with GELU activation. Cross-level 
attention disambiguates patches whose editing evidence is strong at one 
semantic scale but weak at another.
The refined multi-level difference features are aggregated into a dense 
per-patch visual evidence representation:
\begin{equation}
    \small
    \mathbf{E}_v^{(0)} = \mathrm{Agg}\!\Bigl(
        \bigl\{\hat{\mathbf{d}}^l\bigr\}_{l\in\mathcal{L}}
    \Bigr) \in \mathbb{R}^{N \times C'},\quad C'=256,
\end{equation}
where $\mathrm{Agg(\cdot)}$ denotes channel-wise concatenation followed by a 
$1\times1$ convolutional projection. This branch answers: \emph{where does 
the semantic content of the image pair differ?}

\subsubsection{Instruction-Grounding Branch}

The instruction-grounding branch supplies a structurally independent stream of evidence distilled from the editing instruction itself. The majority of instructions specify a well-defined semantic transformation, characterized by an original concept $c^o$, an edited concept $c^e$, and an operation type $op$ (representative examples are provided in Appendix~\ref{app:instruction_examples}). 
We exploit the informative content contained in the instruction by employing a sequence of three consecutive submodules.
 
\noindent\textbf{Semantic Transformation Parser (STP).}
STP extracts the tuple $(c^o, c^e, op)$ from $T$, with $op\!\in\!\{\textit{add},\textit{remove},\textit{replace},\textit{attribute change},\textit{global}\}$. We use Qwen2.5-0.5B~\cite{qwen_Qwen25TechnicalReport_2025} with in-context prompting (full prompt is presented in Appendix~\ref{sec:stp_prompt}), chosen for its strong instruction-following at minimal parameter cost.
 
\noindent\textbf{SAM-guided Attention Construction.}
Each non-empty concept is associated with its corresponding image, $c^o$ with $I^o$ and $c^e$ with $I^e$, and provided to LangSAM (GroundingDINO\cite{liu2024grounding} + SAM2.1\cite{ravi2024sam}) as a textual input prompt in order to derive a concept-conditioned attention map:
\begin{equation}
\phi(I,c)=
\begin{cases}
\small
\epsilon_{\text{global}}\mathbf{1},
    & c=\varnothing \text{ or }  \mathcal{L}(I,c)=\varnothing,\\[2mm]
\mathcal{R}(\max\mathcal{L}(I,c)),
    & \text{otherwise},
\end{cases}
\label{eq:phi}
\end{equation}

where $\mathcal{L}(I,c)$ is the LangSAM prediction, $\max\mathcal{L}$ denotes instance-wise max merging, $\mathcal{R}$ is bilinear resizing to the working resolution. We set the global smoothing parameter to $\epsilon_{\text{global}} = 0.001$ in order to maintain numerical stability in the generation process and to avoid obtaining a completely empty output. We obtain the two downstream concept maps as $A^{o}\!=\!\phi(I^o,c^{o})$ and $A^{e}\!=\!\phi(I^e,c^{e})$.
 
\noindent\textbf{Action-Conditioned Spatial Fusion (ACSF).}
Different actions imply structurally different spatial relationships between $A^o$ and $A^e$, which we encode by an action-conditioned rule:
\begin{equation}
    \small
    M_\mathrm{intent}(p) =
    \begin{cases}
        A^e(p)                  & op = \textit{add},    \\[3pt]
        A^o(p)                  & op = \textit{remove}, \\[3pt]
        A^o(p)\cdot A^e(p)      & op \in \{\textit{replace},\,\textit{attribute change}\}, \\[3pt]
        \epsilon_{\text{global}}\mathbf{1} & op = \textit{global}.
    \end{cases}
    \label{eq:acsf}
\end{equation}
The product rule for \textit{replace} and \textit{attribute change} reflects the joint requirement that patch $p$ shows the original concept in $I^o$ and the edited concept in $I^e$. For the \textit{global} case, the near-zero uniform map provides only weak instruction evidence, naturally letting the semantic branch dominate. Representative visualizations of the instruction-grounding branch are provided in the Appendix~\ref{sec:ActionCondition}.
The intent map is then passed through a $1\!\times\!1$ convolution with ReLU to yield the initial instruction evidence $\mathbf{E}_i^{(0)} = \mathrm{Conv}(M_\mathrm{intent}) \in \mathbb{R}^{N \times C'}$.

\subsubsection{Bidirectional Cross-Modal Refinement}
\label{sec:icmv}
 
The two branches play distinct but coupled roles: the semantic-difference branch acts as the engine, generating dense cues that drive mask prediction, while the instruction-grounding branch acts as the navigator, providing a spatial prior that predicts where edits should occur. To fuse them, the Bidirectional Cross-Modal Refinement (BCMR) module stacks cross-attention blocks with an asymmetric order: at each layer, the semantic-difference branch first attends to the current prior to amplify edit-relevant signals, then the instruction branch attends to this freshly updated state to re-center the prior on regions with strongest visual support. This ordering enforces the intended asymmetry: the semantic-difference branch carries the primary signal, while the instruction prior continuously self-refines to sharpen where the model should act.
 
Formally, let $\mathbf{E}_v^{(0)}$ and $\mathbf{E}_i^{(0)}$ be the initial visual and instruction representations. At each iteration $k=1,\ldots,K$, we perform two sequential cross-attention updates:
\begin{equation}
    \small
    \mathbf{E}_v^{(k)} = \mathbf{E}_v^{(k-1)} + \mathrm{CrossAttn}\bigl(\mathbf{E}_v^{(k-1)}, \mathbf{E}_i^{(k-1)}\bigr), \quad
    \mathbf{E}_i^{(k)} = \mathbf{E}_i^{(k-1)} + \mathrm{CrossAttn}\bigl(\mathbf{E}_i^{(k-1)}, \mathbf{E}_v^{(k)}\bigr).
    \label{eq:bcmr}
\end{equation}
The instruction update queries the freshly updated $\mathbf{E}_v^{(k)}$ rather than the stale $\mathbf{E}_v^{(k-1)}$, so the prior is always re-anchored against the most current visual evidence.

\noindent\textbf{Consensus representation.}
After $K$ iterations, the two refined representations are fused into a shared consensus space:
\begin{equation}
    \small
    \mathbf{Z} = \mathrm{Proj}\!\left(
        \bigl[\mathbf{E}_v^{(K)} \;\|\; \mathbf{E}_i^{(K)}\bigr]
    \right) \in \mathbb{R}^{N \times C'},
    \label{eq:consensus}
\end{equation}
where $\mathrm{Proj}\colon\mathbb{R}^{2C'}\!\to\!\mathbb{R}^{C'}$ is a $1\!\times\!1$ convolution. $\mathbf{Z}$ encodes the image patches at which the two evidence sources intersect, corresponding exactly to the regions that underwent semantic modification.
 
\subsubsection{Mask Decoder}

The mask decoder aggregates DRB outputs $\{\hat{\mathbf{d}}^{\ell}\}_{\ell\in\mathcal{L}}$ with a top-down FPN and then predicts dense mask logits. Although these tensors come from different ViT depths, they are all defined on the same patch lattice $(H_p,W_p)$; the decoder therefore builds a spatial pyramid internally by adaptive pooling of $\hat{\mathbf{d}}^{3}$ and $\hat{\mathbf{d}}^{2}$ to approximately $(H_p/4,W_p/4)$ and $(H_p/2,W_p/2)$ while keeping $\hat{\mathbf{d}}^{1}$ at $(H_p,W_p)$, followed by a $1{\times}1$ projections, top-down fusion, and $3{\times}3$ convolutional layer. 
The highest-resolution FPN feature map is upsampled to the full input size $(H,W)$ and concatenated with $\mathbf{Z}$ (also upsampled to $(H,W)$). A convolutional residual fusion block then distills this combined signal into $\mathbf{\hat{Z}}$. Starting from $\mathbf{\hat{Z}}$, a terminal convolutional residual block diverges into a mask head, which outputs logits corresponding to the edited regions.

\subsection{Noise-Calibrated Progressive Domain Adaptation}
\label{sec:training}

A central bottleneck in text-driven IML is the absence of pixel-accurate ground truth, which causes models trained only on inpainting data to conflate stochastic diffusion noise with semantic change. We address this with a two-stage strategy. Let $f(I^o,I^e)\in[0,1]^{H\times W}$ denote the mask probabilities produced by SIGMA, obtained by applying a sigmoid to the mask-head logits. For an inpainting pair, $M^{\mathrm{inp}}\in\{0,1\}^{H\times W}$ is the pixel-level ground-truth mask of the edited region.

\textbf{Stage~I} pre-trains SIGMA on inpainting pairs with these masks. The segmentation objective adds binary cross-entropy, which scores each pixel independently, to a soft Dice loss~\cite{milletari2016vnet}, which scores region overlap and remains informative when the edited area is small:
\begin{equation}
\small
\begin{aligned}
\mathrm{BCE}(\hat{M},M)
&= -\frac{1}{HW}\sum_{x,y}\Big[M_{xy}\log \hat{M}_{xy}+(1-M_{xy})\log(1-\hat{M}_{xy})\Big],\\
\mathrm{Dice}(\hat{M},M)
&= 1-\frac{2\sum_{x,y}\hat{M}_{xy}M_{xy}+\epsilon}{\sum_{x,y}\hat{M}_{xy}+\sum_{x,y}M_{xy}+\epsilon},
\end{aligned}
\label{eq:bce_dice}
\end{equation}
where $\hat{M}=f(I^o_{\mathrm{inp}},I^e_{\mathrm{inp}})$, $M=M^{\mathrm{inp}}$, and $\epsilon$ is a small constant that keeps the Dice ratio finite. Both terms take a probability map and a binary target and return a scalar loss. We set $\mathcal{L}_\mathrm{seg}=\mathrm{BCE}(\hat{M},M)+\mathrm{Dice}(\hat{M},M)$. Stage~I therefore supplies a strong prior for localized-edit detection, while leaving the model overly sensitive to global pixel perturbations.

\textbf{Stage~II} adapts SIGMA to unlabeled text-driven pairs from public corpora (AnyEdit~\cite{yu_AnyEditMasteringUnified_2025}, CrispEdit-2M~\cite{chow_EditMGTUnleashingPotentials_2026}, OmniEdit-Filtered-1.2M~\cite{wei_OmniEditBuildingImage_2025}, PromptfixData~\cite{yu_PromptFixYouPrompt_2024}) with three self-supervised signals on top of the retained $\mathcal{L}_\mathrm{seg}$.

\textbf{(i) Noise calibration.} For each $I^o$, a VAE-roundtrip $\hat{I}^o=\mathcal{V}_\mathrm{dec}(\mathcal{V}_\mathrm{enc}(I^o)+\epsilon)$ ($\epsilon\!\sim\!\mathcal{N}(0,\sigma^2\mathbf{I})$, $\sigma\!\sim\!\mathcal{U}(0,0.08)$) yields a semantically identical reference whose ground truth is all-zero, supervised by $\mathcal{L}_\mathrm{calib}=\mathrm{BCE}(f(I^o,\hat{I}^o),\mathbf{0})$. The VAE $\mathcal{V}$ is that of Stable Diffusion 1.5~\cite{rombach2022high} or Stable Diffusion XL~\cite{podell2024sdxl}.

\textbf{(ii) EMA self-training.} After each student update, the teacher parameters follow
\begin{equation}
\small
\theta_{\mathrm{EMA}} \leftarrow \mu\,\theta_{\mathrm{EMA}} + (1-\mu)\,\theta,
\label{eq:ema}
\end{equation}
where $\theta$ denotes the trainable student parameters, $\theta_{\mathrm{EMA}}$ denotes the teacher parameters, and the decay is $\mu=0.999$. The teacher $f_\theta^\mathrm{EMA}$ supplies confident pseudo-labels $\hat{M}$ on $(I^o,I^e)$, used by $\mathcal{L}_\mathrm{pl}=C(\hat{M})\cdot\mathrm{BCE}(f(I^o,I^e),\mathrm{sg}(\hat{M}))$ with $C(\hat{M})=\mathbf{1}[\hat{M}>0.8]+\mathbf{1}[\hat{M}<0.2]$. Here $\mathrm{sg}(\cdot)$ is the stop-gradient operator, so the teacher is not updated through $\mathcal{L}_\mathrm{pl}$.

\textbf{(iii) Edit-noise disentanglement.} On the predicted-edit support $P$, we contrast the DRB difference vectors of the real edit pair and the noise pair via $\mathcal{L}_\mathrm{disent}=\frac{1}{|P|}\sum_{i\in\Omega_p} P_i\max(0,\langle\mathbf{d}_\mathrm{edit}^{(i)},\mathbf{d}_\mathrm{noise}^{(i)}\rangle)$, pushing them into separable feature subspaces.

The full objective $\mathcal{L}_\mathrm{Stage\,II}=10\mathcal{L}_\mathrm{seg}+0.1\mathcal{L}_\mathrm{calib}+0.5\mathcal{L}_\mathrm{pl}+0.5\mathcal{L}_\mathrm{disent}$ assigns the dominant weight to the only strongly supervised term while preventing collapse on the self-supervised streams.


\section{Experiments and Analysis}
\label{sec:experiments}

\subsection{Experimental Setup}
\label{sec:setup}

\textbf{Datasets.}
We use three groups of datasets: (i) For evaluation, we adopt five text-driven forgery localization benchmarks that provide original images, edited images, and pixel-level ground-truth masks: CoCoGlide~\cite{guillaro_TruForLeveragingAllRound_2023}, AutoSplice~\cite{jia_AutoSpliceTextPromptManipulated_2023}, MagicBrush~\cite{zhang_MagicBrushManuallyAnnotated_2023}, DEAL-300K~\cite{zhang_DEAL300KDiffusionbasedEditing_2025a}, and OpenSDI~\cite{wang_OpenSDISpottingDiffusionGenerated_2025}, where all benchmarks except OpenSDI provide explicit text instructions describing the edits.They are used both to compare the mask quality of SIGMA against existing automatic generators and as held-out test sets for evaluating the cross-dataset generalization of IML detectors fine-tuned on SIGMA-annotated data. (ii) To generate large-scale training IML data, we employ SIGMA on four extensive public image-editing corpora that were originally distributed without pixel-level annotation masks: AnyEdit~\cite{yu_AnyEditMasteringUnified_2025}, CrispEdit-2M~\cite{chow_EditMGTUnleashingPotentials_2026}, OmniEdit-Filtered-1.2M~\cite{wei_OmniEditBuildingImage_2025}, and PromptfixData~\cite{yu_PromptFixYouPrompt_2024}. Together, these corpora comprise millions of \textit{(original, edited, instruction)} triplets across a broad range of editors and edit types. (iii) For Stage I training of SIGMA, we utilize the BR-Gen\cite{cai_ZoomingFakesNovel_2026} dataset, specifically concentrating on image inpainting operations.

\noindent\textbf{Comparative methodologies for mask generation.}
We compare SIGMA against two families of baselines: (i) pixel-differencing variants without learning, including \textbf{PixDiff-Fixed($\tau$)} with $\tau\in\{0,10,20,30,40\}$, \textbf{PixDiff-Fixed-morph} (on the PixDiff-Fixed binary mask at $\tau{=}25$, we apply morphological opening (5$\times$5 kernel) followed by closing), and \textbf{PixDiff-Otsu} (automatic thresholding); and (ii) learning-based change detectors adapted from remote sensing, namely \textbf{DDPM-CD}~\cite{chamindabandara_DDPMCDDenoisingDiffusion_2025} and \textbf{Meta-CD}~\cite{gao_CombiningSAMLimited_2025}. For fair comparison, the two learning-based baselines are trained on the same inpainting dataset used in SIGMA's Stage~I. 

\begin{table*}[t]
    \centering
    \caption{Quantitative comparison across different datasets. F1 and IoU scores (\%) are reported.}
    \label{tab:maskQuality}
    \small
    \resizebox{\textwidth}{!}{
    \begin{tabular}{l|cc|cc|cc|cc|cc|cc}
    \toprule
    \multirow{2.5}{*}{Methods} & \multicolumn{12}{c}{Datasets} \\
    \cmidrule{2-13}
     & \multicolumn{2}{c}{CoCoGlide} & \multicolumn{2}{c}{AutoSplice} & \multicolumn{2}{c}{MagicBrush} & \multicolumn{2}{c}{DEAL-300K} & \multicolumn{2}{c}{OpenSDI} & \multicolumn{2}{c}{Average} \\
     & F1 & IOU & F1 & IOU & F1 & IOU & F1 & IOU & F1 & IOU & F1 & IOU \\
    \midrule
    PixDiff-Fixed($\tau=0$) & 35.46 & 25.22 & 53.90 & 40.94 & 23.48 & 14.79 & 16.38 & 10.33 & 24.65 & 16.17 & 30.77 & 21.49 \\
    PixDiff-Fixed($\tau=10$) & 92.26 & 86.22 & 92.18 & 86.03 & 74.89 & 62.93 & 28.04 & 19.11 & 68.11 & 56.18 & 71.10 & 62.09 \\
    PixDiff-Fixed($\tau=20$) & 82.01 & 71.00 & 86.15 & 76.57 & 69.11 & 55.52 & 33.44 & 23.28 & 63.85 & 49.65 & 66.91 & 55.20 \\
    PixDiff-Fixed($\tau=30$) & 72.27 & 58.62 & 79.76 & 67.50 & 62.10 & 47.66 & 35.24 & 24.41 & 55.50 & 40.61 & 60.97 & 47.76 \\
    PixDiff-Fixed($\tau=40$) & 63.34 & 48.59 & 73.46 & 59.37 & 55.45 & 40.83 & 35.24 & 24.10 & 47.23 & 32.81 & 54.94 & 41.14 \\
    PixDiff-Fixed-morph & 59.92 & 46.12 & 77.67 & 65.22 & 64.06 & 50.03 & 46.33 & 34.44 & 51.91 & 37.45 & 59.98 & 46.65 \\
    PixDiff-Otsu & 62.40 & 46.54 & 62.97 & 47.07 & 51.88 & 36.84 & 34.21 & 23.17 & 50.07 & 34.95 & 52.31 & 37.71 \\
    Qwen2.5VL-LangSAM & 84.54 & 77.66 & 86.19 & 80.30 & 38.95 & 29.09 & 43.13 & 33.68 & 43.71 & 36.19 & 59.31 & 51.38 \\
    Meta-CD (TGRS 2025) & 73.50 & 62.40 & 86.90 & 77.80 & 76.00 & 64.10 & 45.80 & 33.80 & 74.28 & 62.19 & 71.30 & 60.06 \\
    DDPM-CD (WACV 2025) & \textbf{97.80} & \textbf{96.00} & 96.20 & 92.80 & \textbf{89.90} & \textbf{83.10} & 20.90 & 13.80 & 83.35 & 76.14 & 77.63 & 72.37 \\
    SIGMA & 95.62 & 91.84 & \textbf{96.69} & \textbf{93.74} & 89.03 & 81.91 & \textbf{76.43} & \textbf{65.31} & \textbf{91.37} & \textbf{84.82} & \textbf{89.83} & \textbf{83.52} \\
    \bottomrule
    \end{tabular}}
\end{table*}

\begin{figure}[t]
    \centering
    \includegraphics[width=\textwidth]{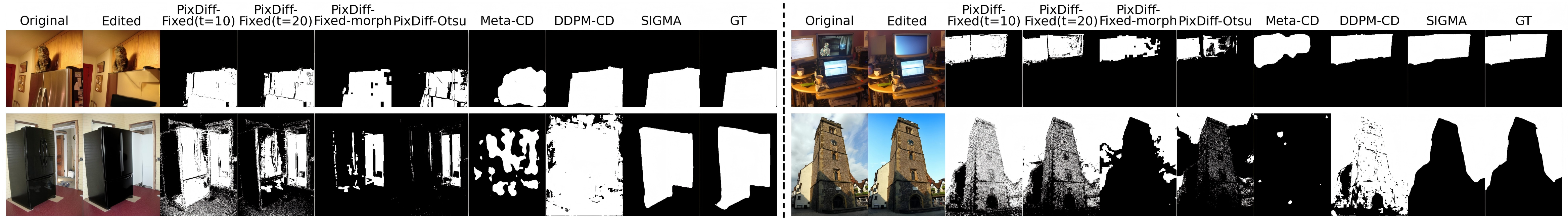}
    \caption{Qualitative comparison of SIGMA and other methods.}
    \label{fig:qualitativeResults}
\end{figure}

\textbf{Implementation details.} We employ DINOv2-Base~\cite{oquab_DINOv2LearningRobust_2024} as a frozen feature-extraction backbone and set the number of DRB modules to 2. The number of iterative refinement steps \(K\) employed in the Bidirectional Cross-Modal Refinement procedure is fixed to 2. The model is optimized using AdamW (with $\beta_1 = 0.9$, $\beta_2 = 0.999$, and weight decay $= 0.05$), an initial learning rate of $10^{-3}$ with cosine annealing, and a batch size of 8 on a single NVIDIA A100 GPU. Input images are resized to $518 \times 518$ pixels. Data augmentation consists of synchronized random horizontal flipping and random cropping, applied identically to both images in each input pair. Flipping does not rewrite side-specific words in the instruction. Appendix~\ref{sec:flipAlign} measures how often this misaligns the instruction with the image and compares a variant that withholds the flip for those samples. 
We trained the SIGMA model for 10 epochs during Stage I and subsequently fine-tuned it for an additional 5 epochs during Stage II.

\subsection{Quality Assessment of Generated Masks}

Tab.~\ref{tab:maskQuality} compares mask quality for pixel-level thresholding, morphological heuristics, and learning-based change detection. PixDiff-Fixed is sensitive to the threshold: at $\tau{=}10$ it reaches 71.10\% average F1, and performance drops for the other values of $\tau$. PixDiff-Otsu reaches 52.31\% F1. DDPM-CD~\cite{chamindabandara_DDPMCDDenoisingDiffusion_2025} reaches 97.80\% F1 on CoCoGlide but only 20.90\% F1 on DEAL-300K, so its 77.63\% average does not hold across datasets. After Stage~II, SIGMA reaches 89.83\% average F1 and 83.52\% IoU, which is 12.20 points above DDPM-CD in F1. The largest margin is on DEAL-300K, where SIGMA exceeds DDPM-CD by 55.53 F1 points. This pattern underscores genuine robustness rather than dataset-specific overfitting. 
Qualitative comparisons (Fig.~\ref{fig:qualitativeResults}) reveal that pixel-difference baselines fail under dense synthesis noise, while Meta-CD and DDPM-CD show complementary failure modes, namely over-dilated masks and missed fine-grained edits, respectively. SIGMA, in contrast, closely matches the ground truth across object-level, identity, and largely edits, visually corroborating its quantitative superiority and reinforcing the necessity of semantic-level, instruction-grounded reasoning. A stratified human audit of these masks on AnyEdit and CrispEdit-2M is reported in Appendix~\ref{sec:humanAudit}.

\subsection{Robustness to Post-Processing Operations}
\label{sec:robustnessAgainstPostProcess}
\begin{figure}[t]
    \centering
    \includegraphics[width=\textwidth]{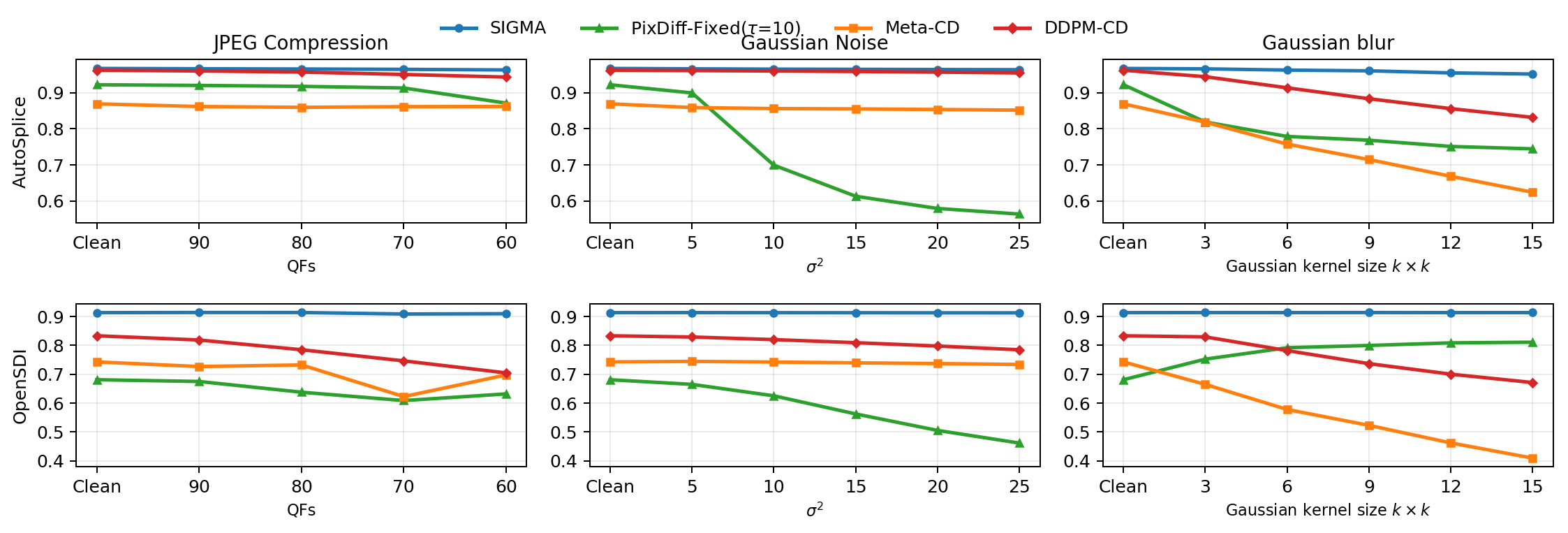}
    \caption{Evaluation of the robustness of SIGMA and comparative baseline methods to JPEG compression, additive white Gaussian noise, and Gaussian blurring on the AutoSplice and OpenSDI datasets.}
    \label{fig:postprocessing}
\end{figure}

To assess the robustness of SIGMA to typical post-processing operations, we subject the test images to three widely used perturbations: JPEG compression (quality factor QF $\in \{90,\ldots,60\}$), additive white Gaussian noise with variance $\sigma^2 \in \{5,\ldots,25\}$, and Gaussian blurring with kernel sizes $\in \{3,\ldots,15\}$. We then quantitatively compare the performance of SIGMA against PixDiff-Fixed ($\tau{=}10$), Meta-CD~\cite{gao_CombiningSAMLimited_2025}, and DDPM-CD~\cite{chamindabandara_DDPMCDDenoisingDiffusion_2025} on the AutoSplice and OpenSDI benchmarks. As shown in Fig.~\ref{fig:postprocessing}, SIGMA delivers near-flat curves across all three perturbation types and severity levels on both benchmarks, holding ${\sim}0.95$ F1 on AutoSplice and ${\sim}0.91$ F1 on OpenSDI throughout. In contrast, PixDiff-Fixed collapses under Gaussian noise (e.g., from ${\sim}0.92$ to ${\sim}0.56$ on AutoSplice at $\sigma^2{=}25$), confirming that pixel-difference signals are intrinsically vulnerable to additive perturbations; Meta-CD and DDPM-CD degrade visibly under blur, with Meta-CD dropping by over 0.24 F1 from clean to a $15{\times}15$ kernel on AutoSplice and from ${\sim}0.74$ to ${\sim}0.41$ on OpenSDI, while DDPM-CD declines by ${\sim}0.13$ on AutoSplice and from ${\sim}0.83$ to ${\sim}0.67$ on OpenSDI under the same kernels. These results confirm that the semantic-feature differencing at the heart of SIGMA confers substantially greater forensic robustness than pixel-level differencing or remote-sensing-style change detection, both of which inherit a strong dependence on the low-level statistics most affected by post-processing.

\begin{table}[t]
  \centering
  \caption{Cross-dataset localization after fine-tuning on automatically annotated editing data. For each detector we report publicly released weights, fine-tuning on the same image pairs annotated by DDPM-CD, and fine-tuning on those pairs annotated by SIGMA (Finetune w/ SIGMA). F1 and IoU scores (\%) are reported; the best result within each method is in \textbf{bold}.}
  \label{tab:LocPerformance}
  \resizebox{\textwidth}{!}{
    \begin{tabular}{llcccccccccccc}
      \toprule
      \multirow{2}{*}{Method} & \multirow{2}{*}{Weights} & \multicolumn{2}{c}{AutoSplice} & \multicolumn{2}{c}{CocoGlide} & \multicolumn{2}{c}{DEAL300K} & \multicolumn{2}{c}{MagicBrush} & \multicolumn{2}{c}{OpenSDI} & \multicolumn{2}{c}{Average} \\
      \cmidrule(lr){3-4} \cmidrule(lr){5-6} \cmidrule(lr){7-8} \cmidrule(lr){9-10} \cmidrule(lr){11-12} \cmidrule(lr){13-14}
      & & F1 & IOU & F1 & IOU & F1 & IOU & F1 & IOU & F1 & IOU & F1 & IOU \\
      \midrule
      \multirow{3}{*}{CAT-Net} & Public Weight & 41.49 & 31.79 & 40.97 & 33.69 & 12.14 & 8.75 & 5.33 & 3.29 & 7.32 & 5.27 & 21.45 & 16.56 \\
      & Finetune w/ DDPM-CD & 44.27 & 30.48 & 32.83 & 21.99 & 13.83 & 8.40 & 59.19 & 42.99 & 20.56 & 12.68 & 34.14 & 23.31 \\
      & Finetune w/ SIGMA & \textbf{60.82} & \textbf{47.23} & \textbf{45.53} & \textbf{34.42} & \textbf{29.99} & \textbf{20.98} & \textbf{61.02} & \textbf{50.05} & \textbf{27.02} & \textbf{18.14} & \textbf{44.88} & \textbf{34.16} \\
      \midrule
      \multirow{3}{*}{MVSSNet} & Public Weight & 18.04 & 12.22 & 32.80 & 25.30 & 5.43 & 3.61 & 1.79 & 1.01 & 1.46 & 0.91 & 11.90 & 8.61 \\
      & Finetune w/ DDPM-CD & 53.98 & 41.03 & 35.51 & 25.27 & 16.40 & 10.34 & \textbf{91.20} & \textbf{85.10} & \textbf{25.06} & \textbf{16.46} & \textbf{44.43} & \textbf{35.64} \\
      & Finetune w/ SIGMA & \textbf{54.14} & \textbf{41.43} & \textbf{35.63} & \textbf{25.38} & \textbf{20.05} & \textbf{13.29} & 81.95 & 73.72 & 21.69 & 14.51 & 42.69 & 33.67 \\
      \midrule
      \multirow{3}{*}{PSCC-Net} & Public Weight & 56.36 & 43.83 & \textbf{50.85} & \textbf{40.60} & 20.19 & 13.69 & 14.87 & 9.37 & 14.50 & 9.64 & 31.35 & 23.43 \\
      & Finetune w/ DDPM-CD & 54.19 & 41.33 & 36.16 & 27.46 & 16.22 & 10.77 & \textbf{41.15} & \textbf{32.71} & 18.54 & 12.10 & 33.25 & 24.87 \\
      & Finetune w/ SIGMA & \textbf{57.06} & \textbf{45.46} & 42.30 & 31.45 & \textbf{21.16} & \textbf{14.20} & 36.69 & 26.67 & \textbf{22.44} & \textbf{15.25} & \textbf{35.93} & \textbf{26.61} \\
      \midrule
      \multirow{3}{*}{Trufor} & Public Weight & 32.69 & 25.52 & 28.33 & 24.33 & 9.26 & 6.77 & 2.51 & 1.48 & 6.93 & 5.22 & 15.94 & 12.66 \\
      & Finetune w/ DDPM-CD & 57.06 & 41.97 & 55.33 & 42.31 & 22.49 & 14.36 & 20.63 & 12.36 & 20.13 & 12.78 & 35.13 & 24.76 \\
      & Finetune w/ SIGMA & \textbf{63.42} & \textbf{50.63} & \textbf{58.72} & \textbf{47.34} & \textbf{27.85} & \textbf{18.92} & \textbf{28.35} & \textbf{20.06} & \textbf{27.02} & \textbf{19.20} & \textbf{41.07} & \textbf{31.23} \\
      \bottomrule
    \end{tabular}
  }
\end{table}

\subsection{Enhancing the Generalization Ability of Existing IML Models}

The principal contribution of SIGMA does not reside in its intrinsic localization accuracy, but rather in its capacity to transform large-scale, unlabeled, text-driven image editing corpora into pixel-level supervisory signals that can be effectively leveraged by arbitrary downstream IML models. We apply SIGMA to large-scale public image-editing datasets to generate corresponding mask annotations (see Appendix~\ref{sec:SIGMA_data} for more details about the SIGMA-annotated dataset), and use the resulting data to fine-tune four representative IML models spanning diverse architectural paradigms: CAT-Net~\cite{kwonCATNetCompressionArtifact2021}, MVSS-Net~\cite{dongMVSSNetMultiViewMultiScale2023a}, PSCC-Net~\cite{liuPSCCNetProgressiveSpatioChannel2022b}, and TruFor~\cite{guillaro_TruForLeveragingAllRound_2023}. These models are finetuned using the implementation from\cite{ma2025imdl}. The fine-tuned models are then evaluated on five disjoint benchmarks: AutoSplice, CoCoGlide, DEAL-300K, MagicBrush, and OpenSDI, to assess their generalization performance. As reported in Tab.~\ref{tab:LocPerformance}, fine-tuning on SIGMA-annotated data raises the average F1 of all four detectors over their public weights, by +23.43\% for CAT-Net (21.45 to 44.88), +30.79\% for MVSS-Net (11.90 to 42.69), +4.58\% for PSCC-Net (31.35 to 35.93), and +25.13\% for TruFor (15.94 to 41.07). On the same image pairs annotated by DDPM-CD, SIGMA gives a higher average F1 for CAT-Net (44.88 versus 34.14), PSCC-Net (35.93 versus 33.25), and TruFor (41.07 versus 35.13). These results show that SIGMA-annotated data is a model-agnostic supervisory resource that lets existing IML detectors use large-scale text-driven editing corpora.

\begin{table*}[t]
\centering
\caption{Ablation study of SIGMA's architectural components and training strategy across five benchmarks. The first six rows progressively add the proposed modules (MSDB, DRBs at varying depths, and IGB), while the last row adds Stage~II adaptation on top of the full architecture. F1 and IoU scores (\%) are reported; best results in each column are in \textbf{bold}.}
\label{tab:ablation}
\resizebox{\textwidth}{!}{
\begin{tabular}{lcccccccccccc}
\toprule
\multirow{3}{*}{Model Variants} & \multicolumn{12}{c}{Datasets} \\
\cmidrule(lr){2-13}
 & \multicolumn{2}{c}{CoCoGlide} & \multicolumn{2}{c}{AutoSplice} & \multicolumn{2}{c}{MagicBrush} & \multicolumn{2}{c}{DEAL-300K} & \multicolumn{2}{c}{OpenSDI} & \multicolumn{2}{c}{Average} \\
 \cmidrule(lr){2-3} \cmidrule(lr){4-5} \cmidrule(lr){6-7} \cmidrule(lr){8-9} \cmidrule(lr){10-11} \cmidrule(lr){12-13}
 & F1 & IOU & F1 & IOU & F1 & IOU & F1 & IOU & F1 & IOU & F1 & IOU \\
\midrule
baseline & 93.80 & 88.90 & 95.49 & 91.79 & 78.34 & 68.00 & 57.89 & 46.22 & 85.54 & 77.13 & 82.21 & 74.41 \\
+msdb & 94.82 & 90.52 & 96.32 & 93.09 & 89.87 & 82.94 & 55.16 & 43.29 & 89.66 & 82.64 & 85.17 & 78.50 \\
+msdb+2DRBs & 95.60 & 91.81 & \textbf{96.82} & \textbf{93.98} & 89.81 & 82.82 & 67.20 & 55.92 & 90.75 & 84.03 & 88.04 & 81.71 \\
+msdb+4DRBs & 95.14 & 90.99 & 96.33 & 93.10 & \textbf{90.04} & \textbf{83.13} & 65.43 & 54.40 & 90.07 & 83.03 & 87.40 & 80.93 \\
+msdb+8DRBs & 95.32 & 91.32 & 96.33 & 93.08 & 89.23 & 82.09 & 68.70 & 57.31 & 90.15 & 83.13 & 87.95 & 81.39 \\
+msdb+2DRBs+IGB & 94.91 & 90.62 & 96.27 & 92.99 & 87.89 & 80.18 & 75.02 & 63.48 & 90.97 & 84.29 & 89.01 & 82.31 \\
+msdb+2DRBs+IGB+Stage II training & \textbf{95.62} & \textbf{91.84} & 96.69 & 93.74 & 89.03 & 81.91 & \textbf{76.43} & \textbf{65.31} & \textbf{91.37} & \textbf{84.82} & \textbf{89.83} & \textbf{83.52} \\
\bottomrule
\end{tabular}%
}
\end{table*}

\subsection{Ablation Studies}
\label{sec:ablation}

\noindent\textbf{Architectural components.} Tab.~\ref{tab:ablation} reports the ablation. Replacing single-scale differencing with the \textbf{Multi-Scale Difference Block (MSDB)} yields +2.96\% F1 and +4.09\% IoU on average, with the largest gains on MagicBrush (+11.53\% F1) and OpenSDI (+4.12\% F1), confirming that edits span heterogeneous scales and benefit from cross-level aggregation. Stacking \textbf{Difference Refinement Blocks (DRBs)} further refines the representation, but the gain is non-monotonic: 2~DRBs reach the best average (88.04\% F1, 81.71\% IoU), while 4 and 8~DRBs slightly degrade easier datasets (e.g., 0.46\% F1 drop on CoCoGlide and 0.49\% on AutoSplice from 2 to 4 DRBs). We trace this to two reinforcing effects: residual stacking steadily smooths the difference map and washes out edit boundaries, while the inpainting-bounded Stage~I supervision starves deeper stacks of useful gradients. Extra parameters thus inject optimization noise instead of real capacity. We therefore adopt 2~DRBs by default. Adding the \textbf{Instruction-Grounding Branch (IGB)} delivers another +0.97\% F1 on average and, most strikingly, raises DEAL-300K F1 from 67.20\% to 75.02\% (+7.82\%), confirming that visual differencing alone cannot disambiguate which synthesis-induced perturbations correspond to user intent and that instruction conditioning is critical for global or semantically nuanced edits.
 
\noindent\textbf{Training strategy.} Stage~I alone attains 89.01\% F1 and 82.31\% IoU, indicating that supervised pre-training on inpainting masks already instills a strong semantic-change prior. However, never having seen diffusion-induced reconstruction noise, it remains partially sensitive to the pervasive pixel perturbations of text-driven edits. Stage~II addresses this by adapting the model to unlabeled editing pairs under three complementary objectives, noise calibration, EMA self-training, and edit-noise feature disentanglement, without requiring any pixel-level annotation in the target domain. The adaptation adds +0.82\% F1 and +1.21\% IoU, with the largest gains on the hardest distributions (+1.41\% F1 on DEAL-300K), and additionally reverses the small regressions that Stage~I alone incurs on CoCoGlide and AutoSplice.

\section{Conclusion and Limitations}
\label{sec:conclusion}

\noindent\textbf{Conclusion.} We presented \textbf{SIGMA}, a framework that turns publicly available image-editing datasets into pixel-accurate supervision for image manipulation localization (IML). Building on the observation that pixel differencing and instruction-only grounding are complementary but individually insufficient under diffusion-based editing, SIGMA performs semantic-feature differencing in a frozen foundation backbone and fuses it with a parsed-instruction prior via bidirectional cross-modal refinement, while a two-stage training strategy bridges supervision scarcity and the diffusion-domain shift. SIGMA outperforms existing automatic mask generators by \textbf{+12.20\%} F1 on five benchmarks and, when applied to public editing corpora, yields a million-scale IML training set that boosts four diverse detectors by \textbf{+20.98\%} F1 in cross-dataset generalization, establishing SIGMA-annotated data as a model-agnostic supervisory resource for the IML community.
 
\noindent\textbf{Limitations.} SIGMA relies on off-the-shelf parsing and grounding models, whose precision limits can occasionally contaminate the instruction prior (mitigated, but not eliminated, by cross-modal refinement). The parser also returns a single operation tuple per instruction. The 1.1M corpus and the benchmarks used here are almost entirely single-round, single-operation edits, and explicitly multi-operation samples account for approximately 0\% of that corpus. Multi-round instructions that apply several operations in sequence are therefore outside the present scope: a later operation can be missed by the instruction branch, and the predicted mask is not split into one region per operation. SIGMA targets semantic edits and is therefore less precise on purely global stylistic changes, where the prompt provides only a weak spatial signal. Finally, our evaluation focuses on still images; extending SIGMA to text-driven video forgeries is an urgent open direction.
 
\noindent\textbf{Broader Impact.} SIGMA aims to strengthen detection of text-driven forgeries and thereby support media verification. The same automated mask-generation capability could in principle refine adversarial editing pipelines. However, SIGMA improves only the detector side rather than the editor, and we believe the net benefit to media integrity is positive.

\begin{ack}
This work was supported in part by the New Generation Artificial Intelligence-National Science and Technology Major Project under Grant 2025ZD0123402, in part by the National Natural Science Foundation of China under Grant 62602719, Grant 62625602, Grant 62441619, Grant 62372489, Grant 62572325, and Grant 6260072840, in part by the Guangdong S\&T Program under Grant 2026B0101100003, in part by the Shenzhen Science and Technology Program under Grant SYSRD20250529113401002 and Grant JCYJ20230807111207015, and in part by Guangdong Provincial Key Laboratory of Information Security Technology.

\end{ack}

\bibliographystyle{unsrt}
\clearpage
\bibliography{ref}

@article{ZhuangDFCN,
  author={Zhuang, Peiyu and Li, Haodong and Tan, Shunquan and Li, Bin and Huang, Jiwu},
  journal={IEEE Transactions on Information Forensics and Security}, 
  title={Image Tampering Localization Using a Dense Fully Convolutional Network}, 
  year={2021},
  volume={16},
  pages={2986-2999},
}

@article{ZhuangReloc,
  author={Zhuang, Peiyu and Li, Haodong and Yang, Rui and Huang, Jiwu},
  journal={IEEE Transactions on Information Forensics and Security}, 
  title={ReLoc: A Restoration-Assisted Framework for Robust Image Tampering Localization}, 
  year={2023},
  volume={18},
  number={},
  pages={5243-5257}
}

@article{li2025towards,
  title={Towards generalizable and robust image tampering localization with multi-task learning and contrastive learning},
  author={Li, Haodong and Zhuang, Peiyu and Su, Yang and Huang, Jiwu},
  journal={Expert Systems with Applications},
  volume={270},
  pages={126492},
  year={2025},
  publisher={Elsevier}
}

@article{yu2026toward,
  title={Toward Real-world Text Image Forgery Localization: Structured and Interpretable Data Synthesis},
  author={Yu, Zeqin and Xie, Haotao and Zhang, Jian and Ni, Jiangqun and Su, Wenkang and Huang, Jiwu},
  journal={Advances in Neural Information Processing Systems},
  volume={38},
  year={2026}
}

@inproceedings{yu2025reinforced,
  title={Reinforced multi-teacher knowledge distillation for efficient general image forgery detection and localization},
  author={Yu, Zeqin and Ni, Jiangqun and Zhang, Jian and Deng, Haoyi and Lin, Yuzhen},
  booktitle={Proceedings of the AAAI Conference on Artificial Intelligence},
  volume={39},
  number={1},
  pages={995--1003},
  year={2025}
}

@inproceedings{yu2023learning,
  title={Learning to locate the text forgery in smartphone screenshots},
  author={Yu, Zeqin and Li, Bin and Lin, Yuzhen and Zeng, Jinhua and Zeng, Jishen},
  booktitle={ICASSP 2023-2023 IEEE International Conference on Acoustics, Speech and Signal Processing (ICASSP)},
  pages={1--5},
  year={2023},
  organization={IEEE}
}

@inproceedings{brooks_InstructPix2PixLearningFollow_2023,
  title = {{{InstructPix2Pix}}: {{Learning To Follow Image Editing Instructions}}},
  shorttitle = {{{InstructPix2Pix}}},
  booktitle = {Proceedings of the {{IEEE}}/{{CVF Conference}} on {{Computer Vision}} and {{Pattern Recognition}}},
  author = {Brooks, Tim and Holynski, Aleksander and Efros, Alexei A.},
  year = 2023,
  pages = {18392--18402}
}

@inproceedings{cai_ZoomingFakesNovel_2026,
  title={Zooming in on fakes: A novel dataset for localized ai-generated image detection with forgery amplification approach},
  author={Cai, Lvpan and Wang, Haowei and Ji, Jiayi and Zhoumen, Yanshu and Chen, Shen and Yao, Taiping and Sun, Xiaoshuai},
  booktitle={Proceedings of the AAAI Conference on Artificial Intelligence},
  volume={40},
  number={4},
  pages={2534--2542},
  year={2026}
}

@inproceedings{chamindabandara_DDPMCDDenoisingDiffusion_2025,
  title = {{{DDPM-CD}}: {{Denoising Diffusion Probabilistic Models}} as {{Feature Extractors}} for {{Remote Sensing Change Detection}}},
  shorttitle = {{{DDPM-CD}}},
  booktitle = {2025 {{IEEE}}/{{CVF Winter Conference}} on {{Applications}} of {{Computer Vision}} ({{WACV}})},
  author = {Chaminda Bandara, Wele Gedara and Nair, Nithin Gopalakrishnan and Patel, Vishal M.},
  year = 2025,
  pages = {5250--5262},
  issn = {2642-9381}
}

@inproceedings{CASIA,
	title={Casia image tampering detection evaluation database},
	author={J. Dong and W. Wang and T. Tan},
	booktitle={IEEE China Summit Inter. Conf. Signal Info. Proc.},
	pages={422--426},
	year={2013},
}

@inproceedings{Columbia,
	title={Detecting image splicing using geometry invariants and camera characteristics consistency},
	author={Y. Hsu and S. Chang},
	booktitle={IEEE Inter. Conf. Multim. Expo},
	pages={549--552},
	year={2006},
	organization={IEEE}
}

@inproceedings{mahfoudi2019defacto,
title={{DEFACTO}: {Image} and Face Manipulation Dataset},
author={Mahfoudi, Ga{\"e}l and Tajini, Badr and Retraint, Florent and Morain-Nicolier, Frederic and Dugelay, Jean Luc and Marc, PIC},
booktitle={{Proceedings of the European Signal Processing Conference}},
pages={1--5},
year={2019},
}

@inproceedings{novozamsky2020imd2020,
  title={{IMD2020}: {A} Large-Scale Annotated Dataset Tailored for Detecting Manipulated Images},
author={Novozamsky, Adam and Mahdian, Babak and Saic, Stanislav},
booktitle={{Proceedings of the IEEE/CVF Winter Conference on Applications of Computer Vision Workshops}},
pages={71--80},
year={2020}
}

@inproceedings{wang_OpenSDISpottingDiffusionGenerated_2025,
  title = {{{OpenSDI}}: {{Spotting Diffusion-Generated Images}} in the {{Open World}}},
  shorttitle = {{{OpenSDI}}},
  booktitle = {Proceedings of the {{IEEE}}/{{CVF Conference}} on {{Computer Vision}} and {{Pattern Recognition}}},
  author = {Wang, Yabin and Huang, Zhiwu and Hong, Xiaopeng},
  year = 2025,
  pages = {4291--4301}
}

@article{dongMVSSNetMultiViewMultiScale2023a,
  title = {{{MVSS-Net}}: {{Multi-View Multi-Scale Supervised Networks}} for {{Image Manipulation Detection}}},
  shorttitle = {{{MVSS-Net}}},
  author = {Dong, Chengbo and Chen, Xinru and Hu, Ruohan and Cao, Juan and Li, Xirong},
  year = 2023,
  month = mar,
  journal = {IEEE Transactions on Pattern Analysis and Machine Intelligence},
  volume = {45},
  number = {3},
  pages = {3539--3553},
  issn = {1939-3539}
}

@inproceedings{kwonCATNetCompressionArtifact2021,
  title = {{{CAT-Net}}: {{Compression Artifact Tracing Network}} for {{Detection}} and {{Localization}} of {{Image Splicing}}},
  shorttitle = {{{CAT-Net}}},
  booktitle = {2021 {{IEEE Winter Conference}} on {{Applications}} of {{Computer Vision}} ({{WACV}})},
  author = {Kwon, Myung-Joon and Yu, In-Jae and Nam, Seung-Hun and Lee, Heung-Kyu},
  year = 2021,
  pages = {375--384},
  issn = {2642-9381}
}

@article{liuPSCCNetProgressiveSpatioChannel2022b,
  title={PSCC-Net: Progressive spatio-channel correlation network for image manipulation detection and localization},
  author={Liu, Xiaohong and Liu, Yaojie and Chen, Jun and Liu, Xiaoming},
  journal={IEEE Transactions on Circuits and Systems for Video Technology},
  volume={32},
  number={11},
  pages={7505--7517},
  year={2022},
  publisher={IEEE}
}

@inproceedings{bandara_TransformerBasedSiameseNetwork_2022,
  title = {A {{Transformer-Based Siamese Network}} for {{Change Detection}}},
  booktitle = {{{IGARSS}} 2022 - 2022 {{IEEE International Geoscience}} and {{Remote Sensing Symposium}}},
  author = {Bandara, Wele Gedara Chaminda and Patel, Vishal M.},
  year = 2022,
  pages = {207--210},
  issn = {2153-7003}
}

@inproceedings{yu_AnyEditMasteringUnified_2025,
  title={Anyedit: Mastering unified high-quality image editing for any idea},
  author={Yu, Qifan and Chow, Wei and Yue, Zhongqi and Pan, Kaihang and Wu, Yang and Wan, Xiaoyang and Li, Juncheng and Tang, Siliang and Zhang, Hanwang and Zhuang, Yueting},
  booktitle={Proceedings of the Computer Vision and Pattern Recognition Conference},
  pages={26125--26135},
  year={2025}
}

@article{gao_CombiningSAMLimited_2025,
  title = {Combining {{SAM With Limited Data}} for {{Change Detection}} in {{Remote Sensing}}},
  author = {Gao, Junyu and Zhang, Da and Wang, Feiyu and Ning, Lichen and Zhao, Zhiyuan and Li, Xuelong},
  year = 2025,
  journal = {IEEE Transactions on Geoscience and Remote Sensing},
  volume = {63},
  pages = {1--11},
  issn = {1558-0644}
}

@article{chow_EditMGTUnleashingPotentials_2026,
  title={EditMGT: Unleashing Potentials of Masked Generative Transformers in Image Editing},
  author={Chow, Wei and Li, Linfeng and Kong, Lingdong and Li, Zefeng and Xu, Qi and Song, Hang and Ye, Tian and Wang, Xian and Bai, Jinbin and Xu, Shilin and others},
  journal={arXiv preprint arXiv:2512.11715},
  year={2025}
}

@article{liu_Step1XEditPracticalFramework_2025,
  title={Step1x-edit: A practical framework for general image editing},
  author={Liu, Shiyu and Han, Yucheng and Xing, Peng and Yin, Fukun and Wang, Rui and Cheng, Wei and Liao, Jiaqi and Wang, Yingming and Fu, Honghao and Han, Chunrui and others},
  journal={arXiv preprint arXiv:2504.17761},
  year={2025}
}

@article{labs_FLUX1KontextFlow_2025,
  title={FLUX. 1 Kontext: Flow Matching for In-Context Image Generation and Editing in Latent Space},
  author={Labs, Black Forest and Batifol, Stephen and Blattmann, Andreas and Boesel, Frederic and Consul, Saksham and Diagne, Cyril and Dockhorn, Tim and English, Jack and English, Zion and Esser, Patrick and others},
  journal={arXiv preprint arXiv:2506.15742},
  year={2025}
}

@article{wu_OmniGen2ExplorationAdvanced_2025,
  title={Omnigen2: Exploration to advanced multimodal generation},
  author={Wu, Chenyuan and Zheng, Pengfei and Yan, Ruiran and Xiao, Shitao and Luo, Xin and Wang, Yueze and Li, Wanli and Jiang, Xiyan and Liu, Yexin and Zhou, Junjie and others},
  journal={arXiv preprint arXiv:2506.18871},
  year={2025}
}

@article{wu_QwenImageTechnicalReport_2025,
  title={Qwen-image technical report},
  author={Wu, Chenfei and Li, Jiahao and Zhou, Jingren and Lin, Junyang and Gao, Kaiyuan and Yan, Kun and Yin, Sheng-ming and Bai, Shuai and Xu, Xiao and Chen, Yilei and others},
  journal={arXiv preprint arXiv:2508.02324},
  year={2025}
}

@article{qwen_Qwen25TechnicalReport_2025,
  title={Qwen2.5 Technical Report},
  author={Qwen An Yang and Baosong Yang and Beichen Zhang and Binyuan Hui and Bo Zheng and Bowen Yu and Chengyuan Li and Dayiheng Liu and Fei Huang and Guanting Dong and Haoran Wei and Huan Lin and Jian Yang and Jianhong Tu and Jianwei Zhang and Jianxin Yang and Jiaxin Yang and Jingren Zhou and Junyang Lin and Kai Dang and Keming Lu and Keqin Bao and Kexin Yang and Le Yu and Mei Li and Mingfeng Xue and Pei Zhang and Qin Zhu and Rui Men and Runji Lin and Tianhao Li and Tingyu Xia and Xingzhang Ren and Xuancheng Ren and Yang Fan and Yang Su and Yi-Chao Zhang and Yunyang Wan and Yuqi Liu and Zeyu Cui and Zhenru Zhang and Zihan Qiu and Shanghaoran Quan and Zekun Wang},
  journal={arXiv preprint arXiv:2412.15115},
  year={2024}
}

@article{oquab_DINOv2LearningRobust_2024,
  title={DINOv2: Learning Robust Visual Features without Supervision},
  author={Oquab, Maxime and Darcet, Timoth{\'e}e and Moutakanni, Th{\'e}o and Vo, Huy and Szafraniec, Marc and Khalidov, Vasil and Fernandez, Pierre and Haziza, Daniel and Massa, Francisco and El-Nouby, Alaaeldin and others},
  journal={Transactions on Machine Learning Research Journal},
  year={2024}
}

@article{yu_PromptFixYouPrompt_2024,
  title={Promptfix: You prompt and we fix the photo},
  author={Yu, Yongsheng and Zeng, Ziyun and Hua, Hang and Fu, Jianlong and Luo, Jiebo},
  journal={Advances in Neural Information Processing Systems},
  volume={37},
  pages={40000--40031},
  year={2024}
}

@inproceedings{
wei_OmniEditBuildingImage_2025,
title={OmniEdit: Building Image Editing Generalist Models Through Specialist Supervision},
author={Cong Wei and Zheyang Xiong and Weiming Ren and Xeron Du and Ge Zhang and Wenhu Chen},
booktitle={The Thirteenth International Conference on Learning Representations},
year={2025},
}

@inproceedings{guillaro_TruForLeveragingAllRound_2023,
  title = {{{TruFor}}: {{Leveraging All-Round Clues}} for {{Trustworthy Image Forgery Detection}} and {{Localization}}},
  shorttitle = {{{TruFor}}},
  booktitle = {Proceedings of the {{IEEE}}/{{CVF Conference}} on {{Computer Vision}} and {{Pattern Recognition}}},
  author = {Guillaro, Fabrizio and Cozzolino, Davide and Sud, Avneesh and Dufour, Nicholas and Verdoliva, Luisa},
  year = 2023,
  pages = {20606--20615}
}

@inproceedings{guo_HierarchicalFineGrainedImage_2023a,
  title = {Hierarchical {{Fine-Grained Image Forgery Detection}} and {{Localization}}},
  booktitle = {Proceedings of the {{IEEE}}/{{CVF Conference}} on {{Computer Vision}} and {{Pattern Recognition}}},
  author = {Guo, Xiao and Liu, Xiaohong and Ren, Zhiyuan and Grosz, Steven and Masi, Iacopo and Liu, Xiaoming},
  year = 2023,
  pages = {3155--3165}
}

@inproceedings{
hertz_PrompttoPromptImageEditing_2022,
title={Prompt-to-Prompt Image Editing with Cross-Attention Control},
author={Amir Hertz and Ron Mokady and Jay Tenenbaum and Kfir Aberman and Yael Pritch and Daniel Cohen-Or},
booktitle={The Eleventh International Conference on Learning Representations},
year={2023}
}

@inproceedings{jia_AutoSpliceTextPromptManipulated_2023,
  title = {{{AutoSplice}}: {{A Text-Prompt Manipulated Image Dataset}} for {{Media Forensics}}},
  shorttitle = {{{AutoSplice}}},
  booktitle = {Proceedings of the {{IEEE}}/{{CVF Conference}} on {{Computer Vision}} and {{Pattern Recognition}}},
  author = {Jia, Shan and Huang, Mingzhen and Zhou, Zhou and Ju, Yan and Cai, Jialing and Lyu, Siwei},
  year = 2023,
  pages = {893--903}
}

@inproceedings{kang_LEGIONLearningGround_2025,
  title={Legion: Learning to ground and explain for synthetic image detection},
  author={Kang, Hengrui and Wen, Siwei and Wen, Zichen and Ye, Junyan and Li, Weijia and Feng, Peilin and Zhou, Baichuan and Wang, Bin and Lin, Dahua and Zhang, Linfeng and others},
  booktitle={Proceedings of the IEEE/CVF International Conference on Computer Vision},
  pages={18937--18947},
  year={2025}
}

@article{su_CanWeGet_2025,
  title = {Can {{We Get Rid}} of {{Handcrafted Feature Extractors}}? {{SparseViT}}: {{Nonsemantics-Centered}}, {{Parameter-Efficient Image Manipulation Localization Through Spare-Coding Transformer}}},
  shorttitle = {Can {{We Get Rid}} of {{Handcrafted Feature Extractors}}?},
  author = {Su, Lei and Ma, Xiaochen and Zhu, Xuekang and Niu, Chaoqun and Lei, Zeyu and Zhou, Ji-Zhe},
  year = 2025,
  journal = {Proceedings of the AAAI Conference on Artificial Intelligence},
  volume = {39},
  number = {7},
  pages = {7024--7032},
  issn = {2374-3468}
}

@inproceedings{sun_ForgerySleuthEmpoweringMultimodal_2025a,
  title = {{{ForgerySleuth}}: {{Empowering Multimodal Large Language Models}} for {{Image Manipulation Detection}}},
  shorttitle = {{{ForgerySleuth}}},
  booktitle = {The {{Thirty-ninth Annual Conference}} on {{Neural Information Processing Systems}}},
  author = {Sun, Zhihao and Jiang, Haoran and Chen, Haoran and Cao, Yixin and Qiu, Xipeng and Wu, Zuxuan and Jiang, Yu-Gang},
  year = 2025
}

@inproceedings{wu_ManTraNetManipulationTracing_2019,
  title = {{{ManTra-Net}}: {{Manipulation Tracing Network}} for {{Detection}} and {{Localization}} of {{Image Forgeries With Anomalous Features}}},
  shorttitle = {{{ManTra-Net}}},
  booktitle = {Proceedings of the {{IEEE}}/{{CVF Conference}} on {{Computer Vision}} and {{Pattern Recognition}}},
  author = {Wu, Yue and AbdAlmageed, Wael and Natarajan, Premkumar},
  year = 2019,
  pages = {9543--9552}
}

@inproceedings{huang2025sida,
  title={Sida: Social media image deepfake detection, localization and explanation with large multimodal model},
  author={Huang, Zhenglin and Hu, Jinwei and Li, Xiangtai and He, Yiwei and Zhao, Xingyu and Peng, Bei and Wu, Baoyuan and Huang, Xiaowei and Cheng, Guangliang},
  booktitle={Proceedings of the Computer Vision and Pattern Recognition Conference},
  pages={28831--28841},
  year={2025}
}

@inproceedings{liu2024grounding,
  title={Grounding dino: Marrying dino with grounded pre-training for open-set object detection},
  author={Liu, Shilong and Zeng, Zhaoyang and Ren, Tianhe and Li, Feng and Zhang, Hao and Yang, Jie and Jiang, Qing and Li, Chunyuan and Yang, Jianwei and Su, Hang and others},
  booktitle={European conference on computer vision},
  pages={38--55},
  year={2024},
  organization={Springer}
}

@article{ravi2024sam,
  title={Sam 2: Segment anything in images and videos},
  author={Ravi, Nikhila and Gabeur, Valentin and Hu, Yuan-Ting and Hu, Ronghang and Ryali, Chaitanya and Ma, Tengyu and Khedr, Haitham and R{\"a}dle, Roman and Rolland, Chloe and Gustafson, Laura and others},
  journal={arXiv preprint arXiv:2408.00714},
  year={2024}
}

@inproceedings{sun2024rethinking,
  title={Rethinking image editing detection in the era of generative ai revolution},
  author={Sun, Zhihao and Fang, Haipeng and Cao, Juan and Zhao, Xinying and Wang, Danding},
  booktitle={Proceedings of the 32nd ACM International Conference on Multimedia},
  pages={3538--3547},
  year={2024}
}

@article{chang2025bytemorph,
  title={Bytemorph: Benchmarking instruction-guided image editing with non-rigid motions},
  author={Chang, Di and Cao, Mingdeng and Shi, Yichun and Liu, Bo and Cai, Shengqu and Zhou, Shijie and Huang, Weilin and Wetzstein, Gordon and Soleymani, Mohammad and Wang, Peng},
  journal={arXiv preprint arXiv:2506.03107},
  year={2025}
}

@inproceedings{
xu2025fakeshield,
title={FakeShield: Explainable Image Forgery Detection and Localization via Multi-modal Large Language Models},
author={Zhipei Xu and Xuanyu Zhang and Runyi Li and Zecheng Tang and Qing Huang and Jian Zhang},
booktitle={The Thirteenth International Conference on Learning Representations},
year={2025},
}

@article{ma2025imdl,
  title={Imdl-benco: A comprehensive benchmark and codebase for image manipulation detection \& localization},
  author={Ma, Xiaochen and Zhu, Xuekang and Su, Lei and Du, Bo and Jiang, Zhuohang and Tong, Bingkui and Lei, Zeyu and Yang, Xinyu and Pun, Chi-Man and Lv, Jiancheng and others},
  journal={Advances in Neural Information Processing Systems},
  volume={37},
  pages={134591--134613},
  year={2025}
}

@inproceedings{yu_DiffForensicsLeveragingDiffusion_2024,
  title = {{{DiffForensics}}: {{Leveraging Diffusion Prior}} to {{Image Forgery Detection}} and {{Localization}}},
  shorttitle = {{{DiffForensics}}},
  booktitle = {Proceedings of the {{IEEE}}/{{CVF Conference}} on {{Computer Vision}} and {{Pattern Recognition}}},
  author = {Yu, Zeqin and Ni, Jiangqun and Lin, Yuzhen and Deng, Haoyi and Li, Bin},
  year = 2024,
  pages = {12765--12774}
}

@article{zhang_DEAL300KDiffusionbasedEditing_2025a,
  title={DEAL-300K: Diffusion-based Editing Area Localization with a 300K-Scale Dataset and Frequency-Prompted Baseline},
  author={Zhang, Rui and Wang, Hongxia and Liu, Hangqing and Zhou, Yang and Zeng, Qiang},
  journal={arXiv preprint arXiv:2511.23377},
  year={2025}
}

@article{zhang_MagicBrushManuallyAnnotated_2023,
  title = {{{MagicBrush}}: {{A Manually Annotated Dataset}} for {{Instruction-Guided Image Editing}}},
  shorttitle = {{{MagicBrush}}},
  author = {Zhang, Kai and Mo, Lingbo and Chen, Wenhu and Sun, Huan and Su, Yu},
  year = 2023,
  journal = {Advances in Neural Information Processing Systems},
  volume = {36},
  pages = {31428--31449}
}

@inproceedings{milletari2016vnet,
  title = {V-Net: Fully Convolutional Neural Networks for Volumetric Medical Image Segmentation},
  author = {Milletari, Fausto and Navab, Nassir and Ahmadi, Seyed-Ahmad},
  booktitle = {2016 Fourth International Conference on 3D Vision (3DV)},
  pages = {565--571},
  year = {2016}
}

@inproceedings{
podell2024sdxl,
title={{SDXL}: Improving Latent Diffusion Models for High-Resolution Image Synthesis},
author={Dustin Podell and Zion English and Kyle Lacey and Andreas Blattmann and Tim Dockhorn and Jonas M{\"u}ller and Joe Penna and Robin Rombach},
booktitle={The Twelfth International Conference on Learning Representations},
year={2024},
}

@inproceedings{rombach2022high,
  title={High-resolution image synthesis with latent diffusion models},
  author={Rombach, Robin and Blattmann, Andreas and Lorenz, Dominik and Esser, Patrick and Ommer, Bj{\"o}rn},
  booktitle={2022 IEEE/CVF conference on computer vision and pattern recognition (CVPR)},
  pages={10674--10685},
  year={2022},
}

\clearpage
\appendix

\section{Scale Disparity between Image Editing and IML Datasets}
\label{sec:dataset_scale}
 
\begin{figure}[thbp]
    \centering
    \includegraphics[width=\textwidth]{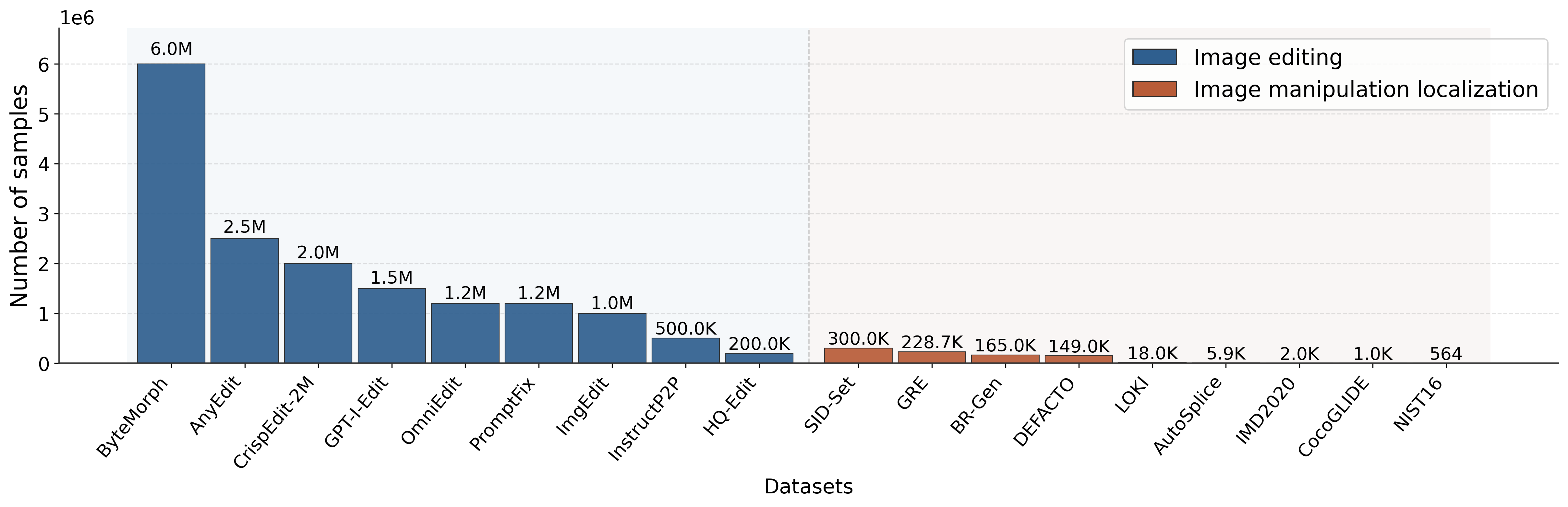}
    \caption{%
    Sample counts of public image-editing datasets (blue) and IML datasets (orange) on a shared linear scale.
    }
    \label{fig:dataset_scale}
\end{figure}

A core motivating premise of this work is the substantial scale asymmetry between publicly available image-editing datasets and image manipulation localization (IML) datasets that supply pixel-accurate masks. We document this asymmetry quantitatively in Fig.~\ref{fig:dataset_scale}, plotting both groups on a shared linear axis to make the gap directly visible. Editing datasets already contain \textit{(original, edited, instruction)} triplets that, by construction, share the manipulation distribution targeted by modern IML; the only missing ingredient is the pixel-level mask. Conversely, IML datasets carry the masks but are limited in scale, in the diversity of editors covered, and in their coverage of the latest text-driven editing paradigms (e.g., FLUX Kontext, OmniGen2, Qwen-Image-Edit).
 
This observation grounds the central question of the paper: rather than collecting another IML benchmark from scratch, can we automatically annotate the editing corpora that already exist? SIGMA is the affirmative answer. By generating pixel-level accurate masks for arbitrary text-driven editing pairs, our approach converts four large-scale datasets (AnyEdit\cite{yu_AnyEditMasteringUnified_2025}, CrispEdit-2M\cite{chow_EditMGTUnleashingPotentials_2026}, OmniEdit-Filtered-1.2M\cite{wei_OmniEditBuildingImage_2025}, and PromptfixData\cite{yu_PromptFixYouPrompt_2024}) to SIGMA-annotated dataset (see Appendix~\ref{sec:SIGMA_data} for more details) into rich supervisory signals that can be leveraged by any downstream IML detector. In doing so, it converts a vast, long-neglected data reservoir into a powerful, model-agnostic IML training pool.

\section{Parsing Editing Instructions into Transformation Tuples}
\label{app:instruction_examples}
Table~\ref{tab:instruction} primarily reports the original and edited concepts embodied in several representative editing instructions, together with their corresponding editing operations. We anticipate that our STP module will efficiently process these instructions, accurately identify the principal conceptual representations, and extract the precise editing operations they specify. 

\begin{table}[thbp]
    \centering
    \renewcommand{\arraystretch}{1.2}
    \caption{Representative editing instructions and the semantic transformation tuples $(c^o, c^e, op)$ extracted by our STP module. $\varnothing$ denotes an empty concept slot, used when an instruction introduces a new concept (\textit{add}), removes an existing one (\textit{remove}), or alters the image globally without a specific subject (\textit{global}).}
    \label{tab:instruction}
    \begin{tabular}{llll}
    \toprule
    Instruction & $c^o$ & $c^e$ & $op$ \\
    \midrule
    ``remove the cat''             & ``a cat''        & $\varnothing$       & remove    \\
    ``add a hat on the person''    & $\varnothing$    & ``a hat''           & add       \\
    ``turn the wood into marble''  & ``wood texture'' & ``marble texture''  & replace   \\
    \makecell[l]{``Change the boy's grey striped \\ t-shirt to a bright blue t-shirt''} 
        & ``grey striped t-shirt'' & ``blue t-shirt'' & attribute change \\
    ``make it more dramatic''      & $\varnothing$    & $\varnothing$       & global    \\
    \bottomrule
    \end{tabular}
\end{table}

\begin{figure}[]
\centering
\newsavebox{\promptbox}
\begin{lrbox}{\promptbox}
\begin{minipage}{\linewidth}
\begin{verbatim}
SYSTEM_PROMPT = """You are an expert at understanding image editing instructions.
Convert each instruction into EXACTLY one JSON object with these keys:
- "original concept"
- "edited concept"
- "action type"

Allowed values for "action type" are only:
1) "add"
2) "remove"
3) "attribute change"
4) "replace"
5) "global"

Definitions:
- add: a new object/concept is introduced.
- remove: an existing object/concept is deleted.
- attribute change: same object identity, only attribute/state/color/pose/
  expression/action/style detail changes.
- replace: one object/concept is swapped with another object/concept.
- global: whole-image change (e.g., weather, season, style, lighting,
  time of day).

Rules:
- Return strict JSON only, no markdown, no extra text.
- If information is implicit, infer concise concepts.
- Keep concepts short noun phrases.

Examples:
Instruction: "add a bird"
{"original concept":"null","edited concept":"a bird","action type":"add"}

Instruction: "remove the umbrella"
{"original concept":"umbrella present","edited concept":"null",
 "action type":"remove"}

Instruction: "change the red flowers to white flowers"
{"original concept":"red flowers","edited concept":"white flowers",
 "action type":"attribute change"}

Instruction: "replace the cat with a dog"
{"original concept":"cat","edited concept":"dog","action type":"replace"}

Instruction: "make it snowy"
{"original concept":"null","edited concept":"null","action type":"global"}
"""
\end{verbatim}
\end{minipage}
\end{lrbox}
\scalebox{0.8}{\usebox{\promptbox}}  
\caption{System prompt used by our Semantic Transformation Parser (STP).}
\label{fig:stp_prompt}
\end{figure}

\section{The Prompt used in Semantic Transformation Parser}
\label{sec:stp_prompt}
Fig.~\ref{fig:stp_prompt} presents the prompt we developed for the Semantic Transformation Parser module. Its primary components comprise task configuration, output format specifications, task definition, rule-based constraints, and contextualized examples. By employing this prompt, we can systematically elicit the large language model’s inference of the tuple $(c^o, c^e, op)$ from a given instruction.

\begin{figure}[t]
    \centering
    \includegraphics[width=\textwidth]{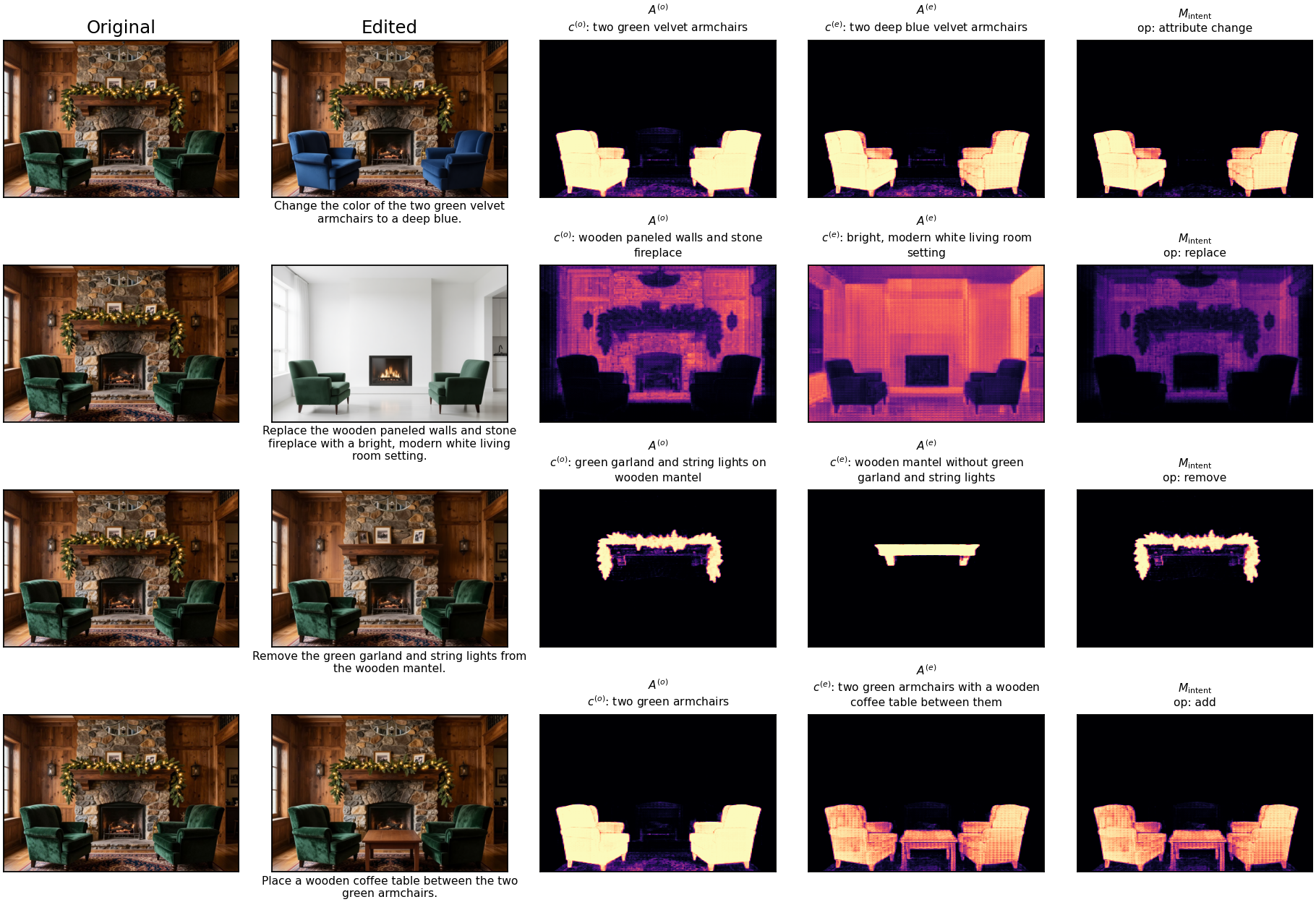}
    \caption{The Visualization of the intermediate image outputs of the text-grounding branch.}
    \label{fig:tgb_visualization}
\end{figure}

\section{Visualization of the Instruction-Grounding Branch}
\label{sec:ActionCondition}

Fig.~\ref{fig:tgb_visualization} visualizes the intermediate outputs of the instruction-grounding branch on four representative editing types, illustrating how the parsed transformation tuple $(c^o, c^e, op)$ propagates through SAM-guided attention construction and action-conditioned spatial fusion. For \emph{attribute change} (row 1, recoloring the armchairs), STP extracts nearly identical concepts that differ only in the color attribute, so $A^o$ and $A^e$ both highlight the armchair regions, and the product rule yields a tightly localized $M_\mathrm{intent}$ that pinpoints the affected objects. For \emph{replace} (row 2, in which the wood-and-stone interior is transformed into a white living room), the original and edited concepts, each grounded in their respective images, span broad and partially overlapping spatial regions. ACSF's product rule preserves only those patches that express the original concept in $I^o$ and the edited concept in $I^e$, specifically the walls and fireplace. This process excludes patches corresponding to unedited furniture. For \emph{remove} (row 3, removing the garland), $c^e$ describes a residual ``mantel without garland''; $A^o$ cleanly localizes the garland on $I^o$, ACSF outputs $A^o$ as the intent map, and the removed region is faithfully recovered.
 
Row~4 (\emph{add}) is informative as a \emph{failure case} of instruction grounding alone: STP over-extracts the contextual phrase ``two green armchairs with a wooden coffee table between them'' as $c^e$, causing $A^e$ to highlight not only the newly added table but also the surrounding armchairs that were never edited. Since the \emph{add} rule directly forwards $A^e$ as $M_\mathrm{intent}$, the resulting prior inherits this contamination, illustrating that the instruction-grounding branch is intrinsically vulnerable to the precision limits of off-the-shelf parsers and grounders. This specific failure mode illustrates why the instruction branch is not designated as authoritative. Within the complete SIGMA pipeline, the bidirectional cross-modal refinement module validates $M_\mathrm{intent}$ against the available semantic evidence. The unaltered armchairs, which exhibit no discernible semantic modifications, are subsequently attenuated or filtered out by the semantic branch during the multimodal fusion stage. The example thus concretely supports our central design decision: semantic differencing and instruction grounding must serve as \emph{mutually verifiable} sources rather than either being trusted alone.

\section{Details of the SIGMA-annotated Dataset}
\label{sec:SIGMA_data}
\begin{figure}[t]
    \centering
    \includegraphics[width=\textwidth]{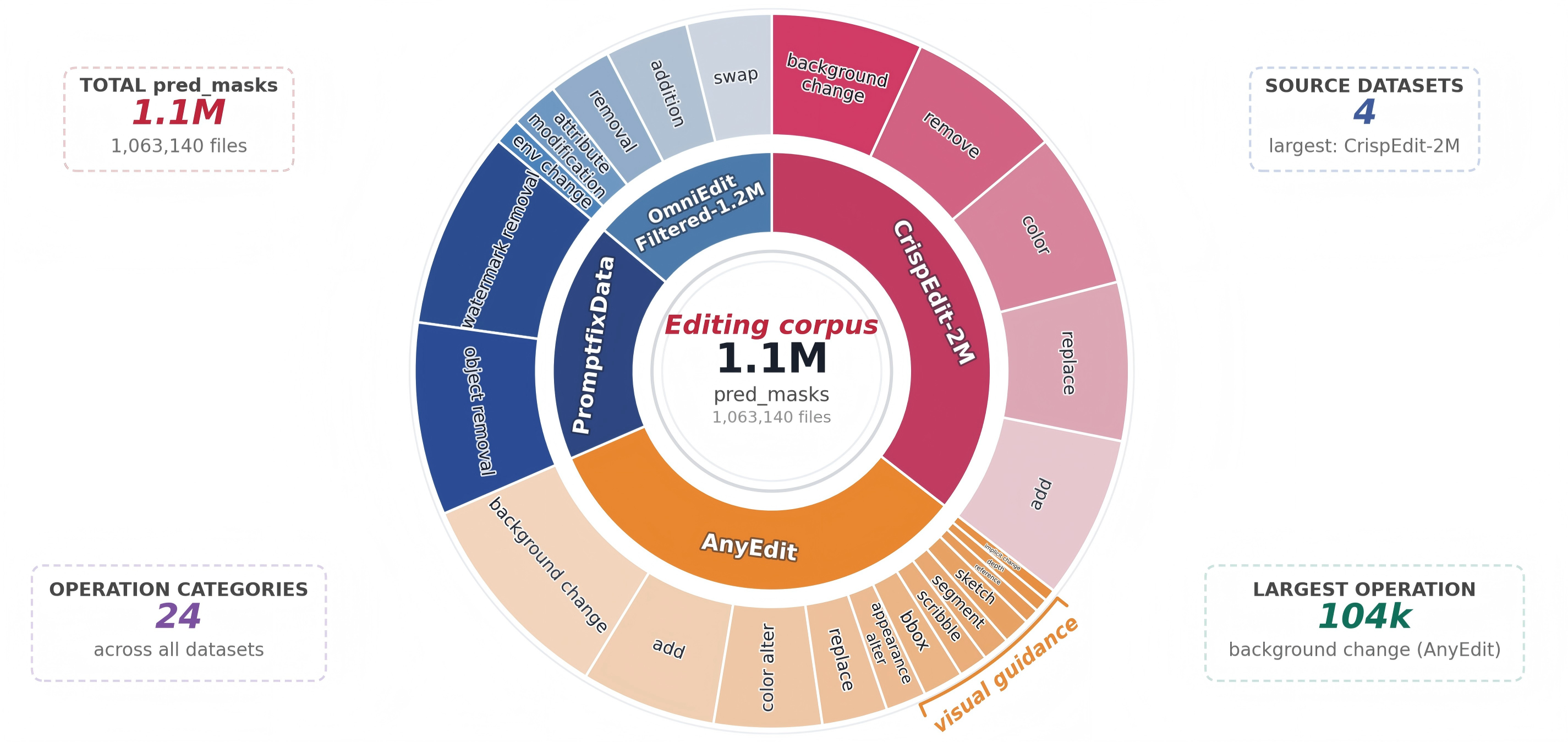}
    \caption{Overview of SIGMA-converted IML training data from four generative image editing corpora and their editing operation taxonomy.}
    \label{fig:convertLocalization}
\end{figure}

We build a large-scale, annotation-rich IML dataset by systematically mining four public generative image editing corpora: AnyEdit\cite{yu_AnyEditMasteringUnified_2025}, CrispEdit-2M\cite{chow_EditMGTUnleashingPotentials_2026}, OmniEdit-Filtered-1.2M\cite{wei_OmniEditBuildingImage_2025}, and PromptfixData\cite{yu_PromptFixYouPrompt_2024}. Each corpus comprises paired original and edited images, accompanied by structured editing metadata, which we transform using SIGMA into a precise, fully automated localization signal, thereby obviating any requirement for manual annotation.

As shown in Fig.~\ref{fig:convertLocalization}, the final dataset contains $\sim$1.1M image pairs covering 24 distinct categories of editing operations, with substantial diversity in both editing semantics and visual appearance. The four sources offer complementary coverage across manipulation regimes. AnyEdit\cite{yu_AnyEditMasteringUnified_2025} provides the broadest semantic coverage to date, comprising 12 carefully curated and specialized operation types. The most frequent categories include large-scale background modification (103K instances), object addition (64K instances), and color transformation (51K instances). The benchmark further incorporates highly localized, brush-level editing primitives (visual scribble, visual segment, and visual sketch), which collectively impose stringent requirements on ultra-fine spatial localization and thus serve as rigorous stress tests for high-precision editing capabilities. CrispEdit-2M\cite{chow_EditMGTUnleashingPotentials_2026} provides the largest single-source corpus (377K samples), characterized by a well-balanced distribution over five object-centric transformation types: addition, background modification, color alteration, object removal, and object replacement, thereby furnishing robust coverage of high-frequency editing intents observed in real-world image manipulation scenarios. OmniEdit-Filtered-1.2M\cite{wei_OmniEditBuildingImage_2025} introduces a set of structurally distinct edit categories, including attribute modification, attribute swapping, and environment transformation. These categories are designed to encompass complex, composite multi-attribute manipulations that present qualitatively different challenges compared to conventional single-operation forgeries. PromptfixData\cite{yu_PromptFixYouPrompt_2024} places a distinctive emphasis on object removal and watermark removal (187K instances), thereby directly addressing the practically important yet historically underexplored problem of content-erasure forgeries. Collectively, these dataset design choices mitigate the dominance of any single editing modality and extend the localization problem to cover a continuum of manipulations, ranging from coarse regional edits to fine-grained, pixel-level erasures.

Ground-truth localization masks are derived automatically from SIGMA as high-fidelity, binary, pixel-level annotations that precisely delineate manipulated regions. Since all source images are generated using contemporary text-guided, diffusion-based editing pipelines, the resulting forgeries exhibit generator-consistent texture statistics, realistic illumination harmonization, and strong semantic coherence. These characteristics distinguish them from traditional cut-and-paste splicing operations and render them representative of the forensic challenges associated with modern AI-generated and AI-edited imagery.

\section{Human Audit on Editing Corpora}
\label{sec:humanAudit}

The five public benchmarks in Tab.~\ref{tab:maskQuality} measure mask quality only where pixel-level ground truth is already available. They leave open whether automatic masks agree with humans on the unlabeled editing corpora that SIGMA is used to annotate. We therefore built a stratified audit from the edit-type distribution in Fig.~\ref{fig:convertLocalization}, drawing 1{,}500 image-instruction pairs from AnyEdit and 1{,}500 from CrispEdit-2M (3{,}000 pairs in total). Two annotators produced a pixel-level change mask for every pair. SIGMA and the competing generators were then scored by pixel-level F1 against these human masks. In addition to the baselines in Sec.~\ref{sec:setup}, the comparison includes the instruction-grounding branch alone (IGB) and Qwen2.5VL-LangSAM, an instruction-only MLLM-plus-grounding pipeline.

SIGMA attains the highest average F1 on both corpora: \textbf{77.35\%} on AnyEdit and \textbf{72.05\%} on CrispEdit-2M (Tab.~\ref{tab:auditAnyEdit} and Tab.~\ref{tab:auditCrisp}). These scores exceed the strongest baseline, PixDiff-Fixed-morph (70.09\% and 60.55\%), by 7.26 and 11.50 points. On AnyEdit, SIGMA is best on Appearance alter (64.59\%), Background change (88.64\%), Color alter (76.48\%), Replace (83.61\%), and four visual-prompt edits (depth, scribble, segment, and sketch), 8 of 11 categories. PixDiff-Fixed($\tau{=}20$) remains higher on Add (74.05\% versus 65.52\%), and PixDiff-Fixed-morph remains higher on Visual bbox (76.57\% versus 75.36\%) and Visual reference (75.42\% versus 68.44\%). On CrispEdit-2M, SIGMA leads every reported category. Its margins over PixDiff-Fixed-morph are largest on Add (+13.48), Background change (+12.21), and Color alter (+14.50). Agreement with humans is therefore strongest on semantic edits whose pixel residual is easily confused with diffusion noise.

The same tables locate the contribution of semantic differencing. IGB alone averages 52.43\% F1 on AnyEdit and 50.19\% on CrispEdit-2M. Its lowest scores are Add (41.35\%), Visual reference (42.92\%), Background change (46.21\%), and Visual scribble (46.42\%) on AnyEdit, and Replace (42.98\%) and Background change (44.69\%) on CrispEdit-2M. Full SIGMA raises the two averages by 24.92 and 21.86 points. Background change rises from 46.21\% to 88.64\% and from 44.69\% to 92.22\%. IGB localizes only the parsed instruction, so these gains are the regions recovered by comparing the image pair directly when parsing or grounding is inaccurate. Pixel F1 summarizes that combined outcome. It does not separate failures of object granularity, prompt ambiguity, parsing, and grounding.

\begin{table*}[t]
\centering
\caption{Pixel-level agreement with human masks on a stratified AnyEdit sample (1{,}500 pairs). F1 scores (\%) are reported. Bold marks the best score in each column.}
\label{tab:auditAnyEdit}
\resizebox{\textwidth}{!}{%
\begin{tabular}{lcccccccccccc}
\toprule
Methods & Add & \makecell{Appearance\\alter} & \makecell{Background\\change} & \makecell{Color\\alter} & Replace & \makecell{Visual\\bbox} & \makecell{Visual\\depth} & \makecell{Visual\\reference} & \makecell{Visual\\scribble} & \makecell{Visual\\segment} & \makecell{Visual\\sketch} & Avg \\
\midrule
PixDiff-Fixed($\tau=0$) & 8.89 & 16.25 & 56.88 & 23.85 & 33.22 & 17.10 & 14.91 & 18.76 & 28.21 & 17.38 & 22.93 & 23.49 \\
PixDiff-Fixed($\tau=10$) & 69.47 & 45.35 & 78.55 & 52.26 & 65.43 & 55.32 & 55.10 & 41.97 & 57.27 & 58.13 & 55.27 & 57.65 \\
PixDiff-Fixed($\tau=20$) & \textbf{74.05} & 46.86 & 75.46 & 55.27 & 73.08 & 71.90 & 72.67 & 54.79 & 72.90 & 73.95 & 68.93 & 67.26 \\
PixDiff-Fixed($\tau=30$) & 73.23 & 42.19 & 71.99 & 48.75 & 73.74 & 75.46 & 75.90 & 61.91 & 74.54 & 75.63 & 70.80 & 67.65 \\
PixDiff-Fixed($\tau=40$) & 71.08 & 36.55 & 68.38 & 40.54 & 71.85 & 73.07 & 73.20 & 65.08 & 71.99 & 72.49 & 68.89 & 64.83 \\
PixDiff-Fixed-morph & 73.42 & 37.10 & 73.16 & 49.53 & 78.64 & \textbf{76.57} & 76.91 & \textbf{75.42} & 76.71 & 78.16 & 75.41 & 70.09 \\
PixDiff-Otsu & 68.98 & 42.53 & 61.28 & 54.54 & 65.33 & 67.38 & 68.39 & 65.22 & 66.40 & 67.91 & 63.99 & 62.90 \\
Meta-CD & 59.60 & 43.85 & 72.31 & 40.80 & 71.70 & 58.13 & 56.16 & 55.28 & 68.35 & 65.77 & 66.93 & 59.90 \\
DDPM-CD & 57.98 & 39.07 & 80.63 & 62.17 & 53.88 & 65.54 & 71.33 & 41.75 & 55.78 & 69.34 & 57.92 & 59.58 \\
IGB & 41.35 & 54.63 & 46.21 & 67.92 & 53.99 & 55.81 & 63.62 & 42.92 & 46.42 & 52.42 & 51.49 & 52.43 \\
Qwen2.5VL-LangSAM & 53.54 & 51.54 & 73.75 & 61.89 & 77.36 & 51.29 & 40.55 & 65.96 & 63.54 & 59.26 & 65.20 & 58.68 \\
SIGMA & 65.52 & \textbf{64.59} & \textbf{88.64} & \textbf{76.48} & \textbf{83.61} & 75.36 & \textbf{77.57} & 68.44 & \textbf{84.59} & \textbf{81.98} & \textbf{84.09} & \textbf{77.35} \\
\bottomrule
\end{tabular}%
}
\end{table*}

\begin{table*}[t]
\centering
\caption{Pixel-level agreement with human masks on a stratified CrispEdit-2M sample (1{,}500 pairs). F1 scores (\%) are reported. Bold marks the best score in each column.}
\label{tab:auditCrisp}
\resizebox{0.92\textwidth}{!}{%
\begin{tabular}{lcccccc}
\toprule
Methods & Add & \makecell{Background\\change} & \makecell{Color\\alter} & Remove & Replace & Avg \\
\midrule
PixDiff-Fixed($\tau=0$) & 25.50 & 70.40 & 22.14 & 29.76 & 30.92 & 35.75 \\
PixDiff-Fixed($\tau=10$) & 46.03 & 78.15 & 46.26 & 42.54 & 41.31 & 50.86 \\
PixDiff-Fixed($\tau=20$) & 52.97 & 79.47 & 55.19 & 47.47 & 43.27 & 55.67 \\
PixDiff-Fixed($\tau=30$) & 55.12 & 78.20 & 58.54 & 48.37 & 42.98 & 56.64 \\
PixDiff-Fixed($\tau=40$) & 55.02 & 75.49 & 59.02 & 47.50 & 41.74 & 55.76 \\
PixDiff-Fixed-morph & 61.51 & 80.01 & 64.30 & 52.54 & 44.40 & 60.55 \\
PixDiff-Otsu & 52.63 & 64.87 & 59.15 & 44.69 & 38.50 & 51.97 \\
Meta-CD & 42.06 & 68.65 & 49.91 & 46.76 & 44.60 & 50.40 \\
DDPM-CD & 45.93 & 77.32 & 48.35 & 45.07 & 39.80 & 51.29 \\
IGB & 54.91 & 44.69 & 58.57 & 49.80 & 42.98 & 50.19 \\
Qwen2.5VL-LangSAM & 55.31 & 73.60 & 54.23 & 46.35 & 50.51 & 59.37 \\
SIGMA & \textbf{74.99} & \textbf{92.22} & \textbf{78.80} & \textbf{59.55} & \textbf{54.68} & \textbf{72.05} \\
\bottomrule
\end{tabular}%
}
\end{table*}

\section{Spatial Alignment under Horizontal Flip}
\label{sec:flipAlign}

The training augmentation in Sec.~\ref{sec:setup} applies the same horizontal flip to both images in a pair and to the mask. The instruction is left unchanged. Words such as ``left'' and ``right'' then point to the opposite side of the flipped image, so the instruction and the visual target can disagree. We checked the four corpora used to build the SIGMA-annotated set, namely AnyEdit, CrispEdit-2M, OmniEdit-Filtered-1.2M, and PromptfixData. Flip-sensitive expressions occur in 5.728\% of the samples. With the 50\% flip probability used in training, about 2.864\% of training samples can receive a flipped image and an instruction that still names the original side.

The corrected policy keeps every other setting fixed and skips horizontal flipping whenever the instruction contains a flip-sensitive expression. We retrain SIGMA under this policy and score it with the same human masks and edit-type split as Appendix~\ref{sec:humanAudit}. Tab.~\ref{tab:flipAnyEdit} and Tab.~\ref{tab:flipCrisp} compare it with the original policy, which flips every pair and keeps the instruction as written. The original policy is the SIGMA row of Tab.~\ref{tab:auditAnyEdit} and Tab.~\ref{tab:auditCrisp}.

Average F1 rises from 77.35\% to 78.66\% on AnyEdit and from 72.05\% to 72.62\% on CrispEdit-2M. The corrected policy is higher in 10 of the 11 AnyEdit categories and in 4 of the 5 CrispEdit-2M categories. The only decreases are on Background change, by 0.79 points on AnyEdit (88.64\% to 87.85\%) and by 0.44 points on CrispEdit-2M (92.22\% to 91.78\%). The gains are modest because the correction changes only the small subset identified above. They are in the same direction on both corpora, so withholding the flip for side-specific instructions removes a genuine mismatch between the image and the instruction without altering the rest of the pipeline.

\begin{table*}[t]
\centering
\caption{Effect of horizontal flip on human-mask agreement for the AnyEdit audit (1{,}500 pairs). F1 scores (\%) are reported. ``Flip all pairs'' applies the original augmentation and keeps the instruction unchanged. ``Skip flip-sensitive'' does not flip samples whose instructions contain side-specific expressions. Bold marks the better score in each column.}
\label{tab:flipAnyEdit}
\resizebox{\textwidth}{!}{%
\begin{tabular}{lcccccccccccc}
\toprule
Policy & Add & \makecell{Appearance\\alter} & \makecell{Background\\change} & \makecell{Color\\alter} & Replace & \makecell{Visual\\bbox} & \makecell{Visual\\depth} & \makecell{Visual\\reference} & \makecell{Visual\\scribble} & \makecell{Visual\\segment} & \makecell{Visual\\sketch} & Avg \\
\midrule
Flip all pairs & 65.52 & 64.59 & \textbf{88.64} & 76.48 & 83.61 & 75.36 & 77.57 & 68.44 & 84.59 & 81.98 & 84.09 & 77.35 \\
Skip flip-sensitive & \textbf{67.66} & \textbf{66.59} & 87.85 & \textbf{76.73} & \textbf{84.77} & \textbf{77.11} & \textbf{78.85} & \textbf{71.29} & \textbf{85.65} & \textbf{83.52} & \textbf{85.27} & \textbf{78.66} \\
\bottomrule
\end{tabular}%
}
\end{table*}

\begin{table*}[t]
\centering
\caption{Effect of horizontal flip on human-mask agreement for the CrispEdit-2M audit (1{,}500 pairs). F1 scores (\%) are reported. The two policies are defined as in Tab.~\ref{tab:flipAnyEdit}. Bold marks the better score in each column.}
\label{tab:flipCrisp}
\resizebox{0.92\textwidth}{!}{%
\begin{tabular}{lcccccc}
\toprule
Policy & Add & \makecell{Background\\change} & \makecell{Color\\alter} & Remove & Replace & Avg \\
\midrule
Flip all pairs & 74.99 & \textbf{92.22} & 78.80 & 59.55 & 54.68 & 72.05 \\
Skip flip-sensitive & \textbf{76.75} & 91.78 & \textbf{79.42} & \textbf{59.88} & \textbf{55.25} & \textbf{72.62} \\
\bottomrule
\end{tabular}%
}
\end{table*}


\end{document}